\documentclass[journal]{IEEEtran}

\pdfoutput=1 
\usepackage{array}
\usepackage{graphicx}
\usepackage{balance} 
\usepackage{url}
\usepackage{float}
\usepackage{amsmath,graphicx,amsfonts,amssymb,url}
\usepackage{setspace}
\usepackage[caption=false]{subfig}
\usepackage{blindtext}
\usepackage[inline]{enumitem}
\usepackage{color}
\usepackage{multirow}

\ifCLASSINFOpdf
\else
\fi
\begin{document}
%
\title{Prediction of the Influence of Navigation Scan-path on Perceived Quality of Free-Viewpoint Videos}

%
%
%
\author{Suiyi~Ling,~\IEEEmembership{Student Member,~IEEE,} Jes{\'u}s~Guti{\'e}rrez, Gu~Ke,~\IEEEmembership{Member,~IEEE,} and~Patrick~Le Callet,~\IEEEmembership{Member,~IEEE}
\thanks{S. Ling, J. Guti{\'e}rrez, and P. Le Callet are with the {\'E}quipe Image, Perception et Interaction, Laboratoire
des Sciences du Num{\'e}rique de Nantes, Universit{\'e} de Nantes, France (e-mail: name.surname@univ-nantes.fr).}
\thanks{G. Ke is with the Faculty of Information Technology, Beijing University of Technology, China (e-mail: guke.doctor@gmail.com).}
\thanks{The work of J. Guti\'errez was supported by the People Programme (Marie Curie Actions) of the European Union`s 7th Framework Programme (FP7/2007-2013) under REA grant agreement n. PCOFUND-GA-2013-609102, through the PRESTIGE Programme coordinated by Campus France.}
\thanks{Manuscript submitted July 31, 2018.}
}


%

\maketitle

\begin{abstract}
Free-Viewpoint Video (FVV) systems allow the viewers to freely change the viewpoints of the scene. In such systems, view synthesis and compression are the two main sources of artifacts influencing the perceived quality. To assess this influence, quality evaluation studies are often carried out using conventional displays and generating predefined navigation trajectories mimicking the possible movement of the viewers when exploring the content. Nevertheless, as different trajectories may lead to different conclusions in terms of visual quality when benchmarking the performance of the systems, methods to identify critical trajectories are needed. This paper aims at exploring the impact of exploration trajectories (defined as Hypothetical Rendering Trajectories: HRT) on perceived quality of FVV subjectively and objectively, providing two main contributions. Firstly, a subjective assessment test including different HRTs was carried out and analyzed. The results demonstrate and quantify the influence of HRT in the perceived quality. Secondly, we propose a new objective video quality assessment measure to objectively predict the impact of HRT. This measure, based on Sketch-Token representation, models how the categories of the contours change spatially and temporally from a higher semantic level. Performance in comparison with existing quality metrics for FVV, highlight promising results for automatic detection of most critical HRTs for the benchmark of immersive systems.

\end{abstract}


\IEEEpeerreviewmaketitle

\section{Introduction}
As immersive multimedia has developed in leaps and bounds along with the emergence of more advanced technologies for capturing, processing and rendering, applications like Free-viewpoint TV (FTV), 3DTV, Virtual Reality (VR) and Augmented Reality (AR) have engaged a lot of users and become the novel hot topic in the multimedia field. In this sense, FTV, which allow the users to immerse themselves into a scene by freely switching the viewpoints as they do in the real world, enables Super Multi-View (SMV) and Free Navigation (FN) applications. On one side, in SMV an horizontal set of more than 80 views (linearly or angularly arranged) is needed to provide users a 3D viewing experience with wide-viewing horizontal parallax, and smooth transition between adjacent views. On the other hand, for FN only a limited set of input views is required, coming from sparse camera arrangements in large baseline setup conditions. In both cases, to deal with such a huge amount of data for delivery and storage, efficient compression techniques are essential, together with robust view-synthesis algorithms, such as Depth-Image-Based Rendering technology (DIBR), which allows to reconstruct the FVV content from a limited set of input views.

\textbf{Importance and difficulties of video quality assessment in FVV:}
video Quality Assessment Metric (VQM) is desirable for evaluating video systems' performance, covering the whole processing chain, from capturing to rendering. In this sense, while hardware developments are leading the advances for capturing and rendering FVV, compression techniques and view synthesis algorithms are main focus of research, as reflected by the ongoing standardization activities within MPEG~\cite{lafruit2015call}\cite{Hinds2017}. This is mainly  due to their importance on the perceived quality, and thus, on the success of the related applications and services~\cite{hanhart2014free}.   

Aside from the well-known compression artifacts, view synthesis techniques (such as Depth Image Based Rendering) have to deal with disoccluded regions~\cite{merkle2009effects}. It is due to the reappearing of the sheltered regions, which are not shown in the reference views but are made visible later in the generated ones. Techniques to recover disocluded regions often introduce geometric distortion and ghosting artifacts. These synthesized-related artifacts are different in nature to compression artifacts, since, they mostly appear locally along the disoccluded areas, while compression artifacts are usually spread over the entire video. In addition, view-synthesis artifacts increase with the baseline distance (i.e., number of synthesized views between two real views) till a point they may be dominant over compression artifacts~\cite{Hinds2017}. Thus, it is very unlikely that VQM proposed for compression-related distortions would be efficient for predicting the quality of sequences produced using synthesized views. 

 
\textbf{Impact of navigations scan-path on perceived quality: free navigation vs predefined trajectories}

Immersive media technologies offer to the users more freedom to explore the content allowing more interactive experiences, than with traditional media. These new possibilities introduce the observers' behavior as an important factor for the perceived quality~\cite{Hinds2017}.

Given the fact that each observer can explore the content differently, there are two approaches can be adopted to practically study this factor: (1) let the observers to navigate the content freely;  (2) let the observer to watch the sequences in form of certain pre-defined navigation trajectories. By employing the first approach, one could obtain a common trajectory according to all the observers' data. However, this common trajectory does not necessarily represent the critical one that will stress the system to the worse case. Moreover, if observers are allowed to navigate freely during the test, it will become a new factor that increase the variability of the mean opinion score (MOS), despite observer's variability in forming quality judgment. As a result, more observers are likely to be required to obtain MOS that can distinguish one system from another statistically significantly.
The second approach (predefined trajectories) is not affected by this trajectory-source of variability but comes with the challenge of selecting the "right" trajectory. In case of system benchmark, one could define "right" trajectory as the most critical one or weakest link, e.g. the one leading to the lowest perceived quality. Nevertheless, there is a good chance that this trajectory-effect is highly dependent on content, some being more sensitive than some others to the choice of trajectory. Identifying the impact of navigation trajectory among different viewpoints on perceived quality for a given content is then of particular interest. For quality evaluation it may be useful to know how navigation affects the visual experience and which are the "worst" trajectories for the system, to carry out performance evaluations of the system under study in the most stressful cases. Consequently, the availability of computational tools to select the critical trajectories would be extremely useful.



\textbf{Contribution}: Based on the discussion above, there are two main research questions in this paper, including (1) does how observer navigate FVV content affect perceived quality; (2) if trajectory affect quality, how to develop a objective metric to indicate "worse" trajectory. To answer these two questions, the contribution of this paper is twofold. Firstly, a subjective test is conducted to study the impact of the exploration trajectory on perceived quality in FVV application scenarios, containing compression and view-synthesis artifacts. In this sense, the concept of Hypothetical Rendering Trajectory (HRT) is introduced. Also, the annotated database obtained from this test is released for research purposes in the field. Secondly, a full-reference Sketch-Token-based synthesized Video Quality Assessment Metric (ST-VQM) is proposed by quantifying to what extent the classes of contours change due to synthesis. This metric is capable of predicting if sequences based on a given trajectory are of higher/lower quality than sequences based on other trajectories, with respect to subjective scores. 

The remainder of the paper is organized as follows. In Section~\ref{sec:RW}, an overview of the state-of-the-art in terms of subjective and objective quality evaluation in relation with FVV scenarios is presented and discussed. In Section~\ref{sec:SE}, the details of the subjective experiment are described, while Section~\ref{sec:OVQA} introduces the proposed VQA metric based on mid-level descriptor. The experimental results from the subjective experiment and the performance evaluation of the proposed objective metric are presented in Section~\ref{sec:sub-ER} and Section~\ref{sec:ER}. Finally, conclusions are given in Section~\ref{sec:con}.


\section{Related Work}\label{sec:RW}
\subsection{Subjective studies}
Although the development of technical aspects related to FTV has been addressed already for some years, the subjective evaluation of the QoE of such systems is still an open issue~\cite{Battisti2016}. As previously mentioned, the majority of the existing studies have been carried out using conventional screens and limiting the interactivity of the users by showing some representative content or predefined trajectories simulating the movement of the observers~\cite{carballeira2015subjective}. In FVV systems, this is especially the case given the limited access to SMV or light-field displays, since only a few prototypes are already available. Nevertheless, it is worth noting the preliminary subjective study that Dricot \textit{et al.}~\cite{Dricot2015} carried out a considering coding and view-synthesis artifacts using a light-field display.   

\begin{figure}[!t]
\subfloat[]{
\includegraphics[width=0.24\textwidth]{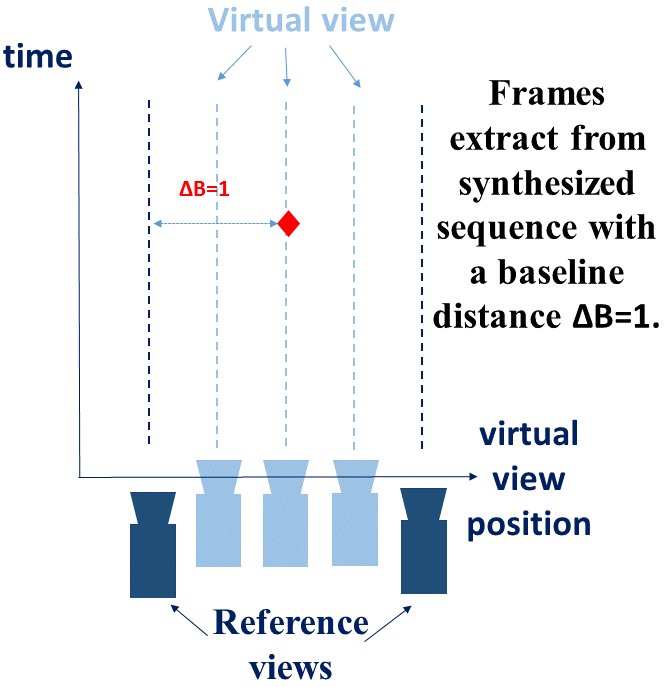}}
\subfloat[]{
\includegraphics[width=0.24\textwidth]{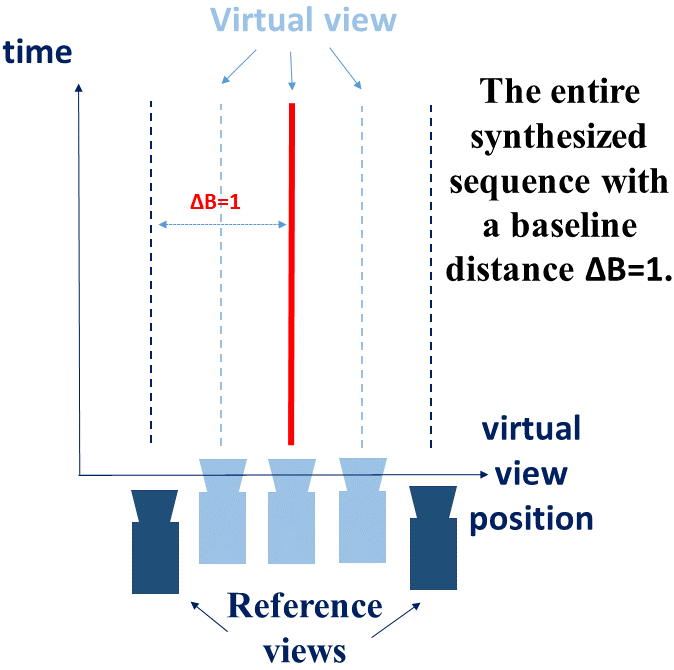}}\\
\subfloat[]{
\includegraphics[width=0.24\textwidth]{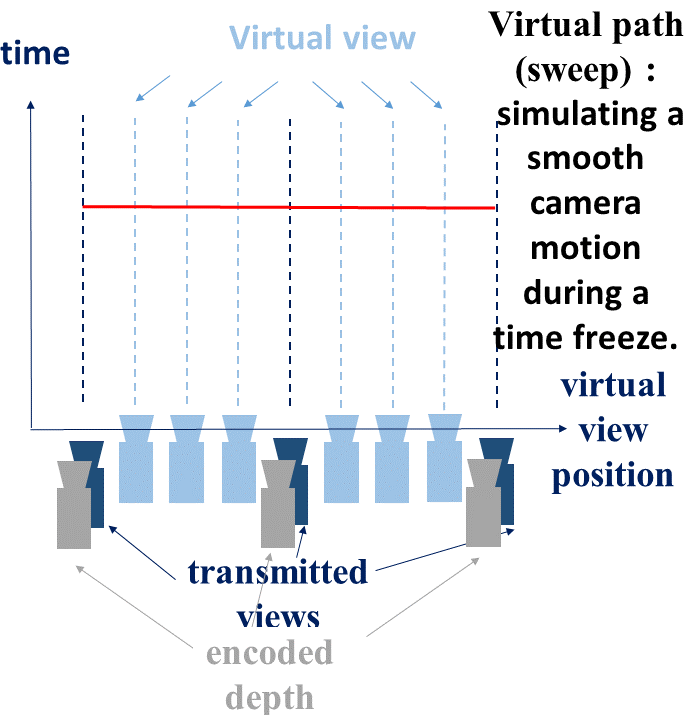}}
\subfloat[]{
\includegraphics[width=0.24\textwidth]{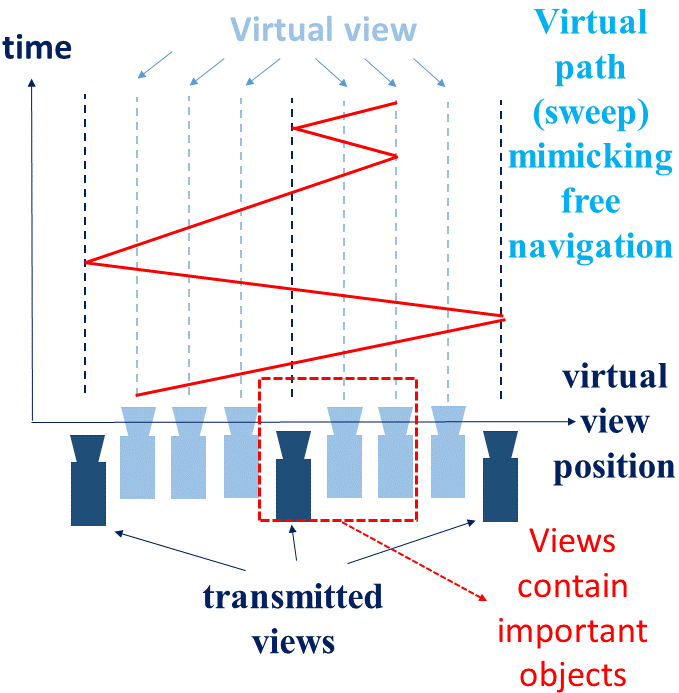}}\\
\caption{Different possibilities to evaluate FTV content representing different degrees of navigation. (a) Synthesized image. (b) Video from a synthesized view (exploration along time). (c) Video containing a view sweep (exploration along views). (d) Video containing a view sweep from videos of various synthesized views (exploration along time and views)  } 
 
\label{fig:sweep}
\end{figure}

In addition to compression techniques, the evaluation and understanding of view-synthesis algorithms is crucial for a successful development of FTV applications and is still an open issue~\cite{Hinds2017}. In this sense, some works that were carried out with previous technologies (e.g., multi-view video), should be taken into account in the study of the effects of view-synthesis in current FTV applications. Firstly, Bosc~\textit{et al.} carried out subjective studies to evaluate the visual quality of synthesized views using DIBR. In these studies, the quality performance of view synthesis was evaluated through different ways, such as: a) the quality of synthesized still images~\cite{bosc2011towards}, b) the quality of videos showing a synthesized view of Multi-View plus Depth (MVD) video sequence~\cite{Bosc2013b}, and c) video sequences showing a smooth sweep across the different viewpoints of an static scene~\cite{bosc2013quality}. These different approaches are represented in Fig.~\ref{fig:sweep}, showing that the first approach only considers spatial DIBR-related artifacts, the second approach considers also temporal distortions within the synthesized view, and the third approach considers spatial DIBR-related artifacts of all the views. To complete the evaluation, another use case should consider the use of view-sweep over the views in video sequences, as depicted in Fig.~\ref{fig:sweep}(d) (i.e., generating videos in which a sweep across the different viewpoints is shown, as if the observer was moving his head horizontally). This approach has been recently adopted in  subjective studies with SMV~\cite{carballeira2015subjective}, which were carried out to study different aspects of this technology, such as smoothness in view transitions and comfortable view-sweep speed~\cite{carballeira2017multiview}, and the impact of coding artifacts~\cite{Recio2017}. MPEG has adopted this type of alternative for their current standardization activities regarding the evaluation of compression techniques for FTV~\cite{lafruit2015call}.

Furthermore, as a result from subjective tests, the availability of appropriate datasets is a crucial aspect for the research on both subjective and objective quality. Especially for supporting the development of objective quality metric, databases containing suitable stimuli (images/videos) annotated with results from subjective are essential. Some efforts have been already made to publish datasets containing free-viewpoint content~\cite{Battisti2016} and some of the aforementioned subjective tests~\cite{Bosc2013b}\cite{bosc2011towards}\cite{bosc2013quality}\cite{liu2015subjective}\cite{Song2015}. Nevertheless, none of these dataset has considered the effect of content adapted trajectories in the "view-sweeping along time" scenario.

\subsection{Objective metrics}
 
Some image quality metrics have been recently proposed especially design to handle view-synthesis artifacts. For instance, Battisti \textit{et al.}~\cite{battisti2015objective} proposed a metric based on statistical features of wavelet sub-bands. Furthermore, considering that using multi-resolution approaches could increase the performance of image quality metrics, Sandi{\'c}-Stankovi{\'c}~\textit{et al.} proposed to use morphological wavelet decomposition~\cite{sandic2015dibr}, and multi-scale decomposition based on morphological pyramids~\cite{sandic2015dibrMP}. Later, the reduced version of these two metrics is presented in~\cite{sandic2016dibr} claiming that PSNR is more consistent with human judgment when calculated at higher morphological decomposition scales.

All the aforementioned metrics are limited to quality assessment of synthesized static images, so they do not explicitly consider temporal distortions that may appear in videos containing synthesized views. Some ad hoc video metrics have been proposed. Zhao and Yu~\cite{zhao2010perceptual} proposes a measure which calculates temporal artifacts that can be perceived by observers in the background regions of the synthesized videos. Similarly, Ekmekcioglu~\textit{et al.}~\cite{ekmekcioglu2010depth} proposed a video quality measure using depth and motion information to take into account where the degradations are located. Moreover, another video metric was recently introduced by Liu~\textit{et al.}~\cite{liu2015subjective} considering the spatio-temporal activity and the temporal flickering that appears in synthesized video sequences. 
However, the aforementioned video quality measures are able to predict the impact of view-synthesis degradations comparing videos corresponding with one single view (as represented in Fig.~\ref{fig:sweep}(b)). In other words, switching among views (resulting from the possible movement of the viewers) and related effects (e.g., inconsistencies among views, geometric flicker along time and view dimensions, etc.~\cite{Battisti2016}) are not addressed. Hanhart~\textit{et al.}~\cite{hanhart2014free} evaluated the performance of state-of-the-art quality measures for 2D video in sequences generated by view-sweep~\cite{bosc2013quality} (as depicted in Fig.~\ref{fig:sweep}(c)), thus considering view-point changes, and reported low performance of all measures in predicting the perceptual quality. An efficient objective video quality measure able to deal with the "view-sweeping along time" scenario is still needed.

\section{Subjective study of the impact of trajectory on perceived quality }\label{sec:SE}
As described in the introduction, the first research question of this paper is to identify the impact of navigation trajectory among different viewpoints on perceived quality taking contents into account. To this end, a subjective study is conducted by designing content related trajectories. A video quality database for FVV scenarios is built, including both compression and view-synthesis artifacts and containing the scores from the subjective assessment test describe in the following. The videos in this database are generated by simulating exploring trajectories that the observers may use in real scenarios, which are set by the Hypothetical Rendering Trajectory (HRT), defined in the following subsection. This database is named as 'Image, Perception and Interaction group Free-viewpoint video  database' (IPI-FVV)\footnote{The public link for downloading the database will be added in the final version of this paper}. 

\subsection{Hypothetical Rendering Trajectory}
A commonly used naming convention for subjective quality assessment studies was provided by the Video Quality Experts Group~\cite{VQEG2010b}, including: SRC (i.e., source or original sequences), HRC (i.e., Hypothetical Reference Circuit or processing applied to the SRC to obtain the test sequences, such as compression techniques), PVS (i.e., Processed Video Sequence or the resulting test sequence from applying an HRC to a SRC). In the context of FN, one should reflect another dimension of the systeme under test related to the interactivity part (e.g. the use of exploration trajectories in quality evaluation of immersive media). Towards this goal, we introduce the term Hypothetical Rendering Trajectories (HRT), to reference the simulated exploration trajectory that is applied to a PVS (as the result of a HRC on a give SRC) for rendering. It is worth mentioning the generality of this term applicable to all immersive media from multiview video to VR, light fields, AR and point clouds.

\subsection{Test Material}
\label{subsec:GTM}

Three different SMV sequences are utilized in our study. These three sequences are Champagne Tower (CT), Pantomime (P) and Big Buck Bunny Flowers (BBBF). Description of the three SMV sequences are summarized in Table \ref{tab:Table 2}. They were also selected as test materials in \cite{lafruit2015call}. For each of the 3 SRC sequences, 20 HRCs, were selected, covering 5 baselines and 4 rate-points (RP). In addition, 2 HRTs were also included to generate 120 PVSs. Details on these parameters, which were selected after a pretest with expert viewers', are described in the following subsections.

\begin{table*}[!ht]\centering \small
\caption{Information of the sequences, including properties and  selected configuration (Rate-point (RP) and baseline distance) }
\label{tab:Table 2}
\begin{tabular}{|c|c|c|c|c|c|c|c|c|c|c|c|}\hline
\multirow{ 2}{*}{Name} &  	\multirow{ 2}{*}{Views} & \multirow{ 2}{*}{ Resolution}  & \multirow{ 2}{*}{ Fps}&	\multirow{ 2}{*}{Seconds} &\multirow{ 2}{*}{ Frames} & \multicolumn{4}{ c |}{QP values}  & \multirow{ 2}{*}{ Baseline Distance} \\ \cline{7-10}
 
 &   &   &   &   &   &  RP1 	&  RP2 	&  RP3 	&  RP4 & \\ \hline
          
BBBF &	91&	1280 x 768&	24&	5  &121 & 35	& - &  45 & 50  &B0, B2, B5, B9, B13\\ \hline
CT	& 80 & 1280 x 960 & 29.4& 10 &300 &  37 	& 43 &	-	&  50 & B0, B4, B8, B12, B16\\ \hline
P & 80 & 1280 x 960 & 29.4 & 10   &300 &	 37  & 43 	&-	& 50  &  B0, B2, B6, B12, B16\\ \hline

\end{tabular}
\end{table*}



\subsubsection{Camera configuration}
\begin{figure}[!htbp]
\begin{center}
 \includegraphics[width=1\columnwidth]{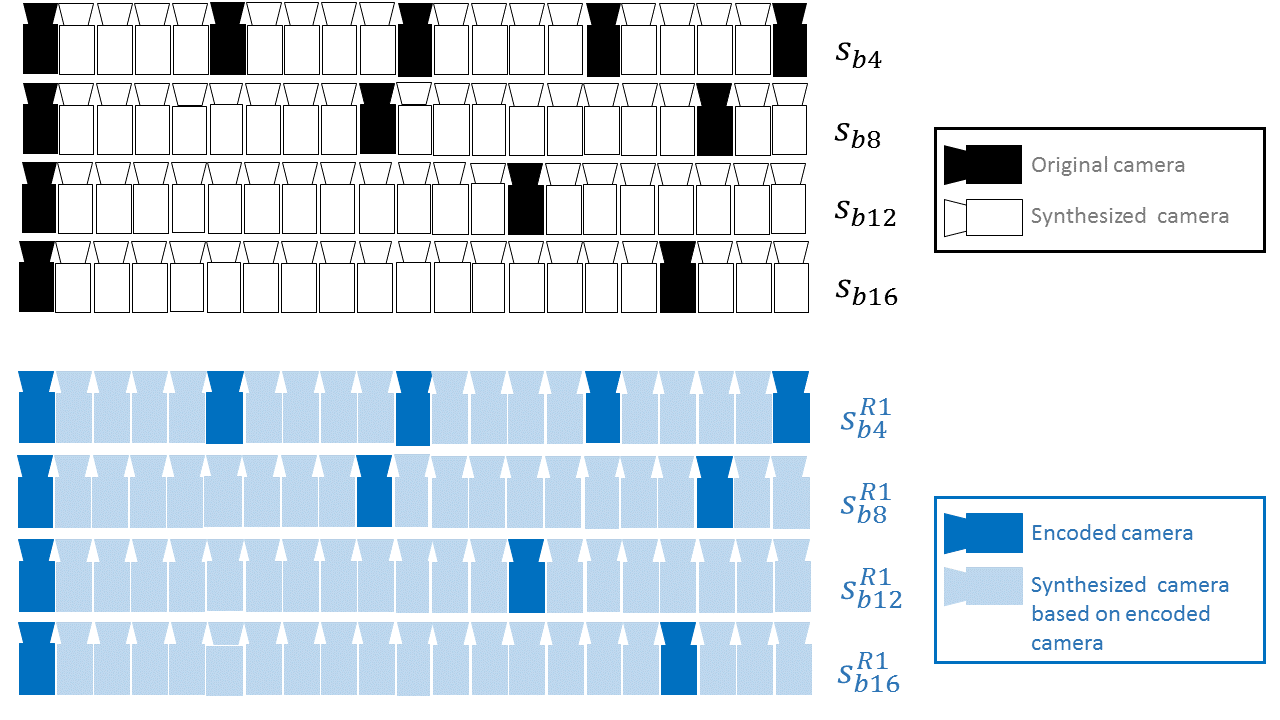}
  \caption{Camera arrangements (1) The upper part of the figure is the configuration designed in \cite{carballeira2015subjective,carballeira2017multiview} where the black cameras represent the sequences taken with real original cameras while the white ones indicate the synthesized view using the original ones as reference (2) The lower part of the figure is the camera  configuration in our experiment, the deep blue camera represents the encoded/transmitted sequence taken from the corresponding original camera while the lighter blue ones indicate the synthesized ones using the encoded ones as reference.       }
  \label{Fig:cam_config}
  \end{center}
\end{figure}

For each source sequence (SRC), 5 stereo baseline values, as summarized in TABLE \ref{tab:Table 2}, are selected in the test including the setting $S_{b_0}$ without using synthesized views.  The baseline is measured based on the camera distances/gaps between left and right real views. Here, $B_i$ or $b_i$ represents the stereo baseline distances that were settled to generate the synthesized virtual views, where $i$ is the number of synthesized views between two reference views. For instance, for camera setting $S_{b4}$  in the the upper part of Fig. \ref{Fig:cam_config}, between each pair of views that captured by original cameras (indicated by two closest black cameras in the figure) there are four virtual views are synthesized using them as left, right  reference. In this case, the baseline distance is 4, denoted as $b_4$. Fig.\ref{Fig:cam_config} illustrates the baseline setting for synthesized views generation in the subjective study. For example, in the lower part of Fig.\ref{Fig:cam_config}, for $S_{b4}^{R1}$, between each two transmitted encoded views, there are totally 4 virtual synthesized views were generated.

\subsubsection{3D-HEVC configuration}
In our experiment, HTM 13.0 in 3D High Efficiency Video Coding  (3D-HEVC) mode was used to encode all the views of the three selected SMV sequences. These encoded views along with the selected original views will be used as the reference views in the following synthesis process, which are also named as 'anchors'. The configuration of the 3D-HEVC encoder recommended in \cite{lafruit2015call} is adopted in the experiment. Specifically, in this experiment, taking into account the contents and the limitations of the duration of subjective experiment tests, 3 rate-points, as summarized in Table~\ref{tab:Table 2}, were selected for each SRC according to the results of the pretest. For each content, the original sequences without compression are also included in the experiment and noted as $RP_0$.

\subsubsection{Depth maps and virtual views generation}
In this paper, reference software tools were used for the preparation of the synthesized views, including Depth Estimation Reference Software (DERS) and View Synthesis Reference Software (VSRS), which have been developed throughout the MPEG-FTV video coding exploration and standardization activities. To generate virtual views with reference sequences taken by real cameras, depth maps and related camera parameters are needed. For sequences 'CT' and 'P'~\cite{NUS}, since original depth maps were not provided, DERS in version of 6.1 is used to generate depth map for each corresponding view. Relative parameters are set as recommended in~\cite{wegner34302ders,lafruit2015ftv}. For synthesized views-generation, the version 4.1 of VSRS is applied. The configuration of the relative parameters is set according to~\cite{lafruit2015ftv} for each corresponding content.

\subsubsection{Navigation trajectory generation}
\begin{figure}[!htbp]
\begin{center}
 \includegraphics[width=1\columnwidth]{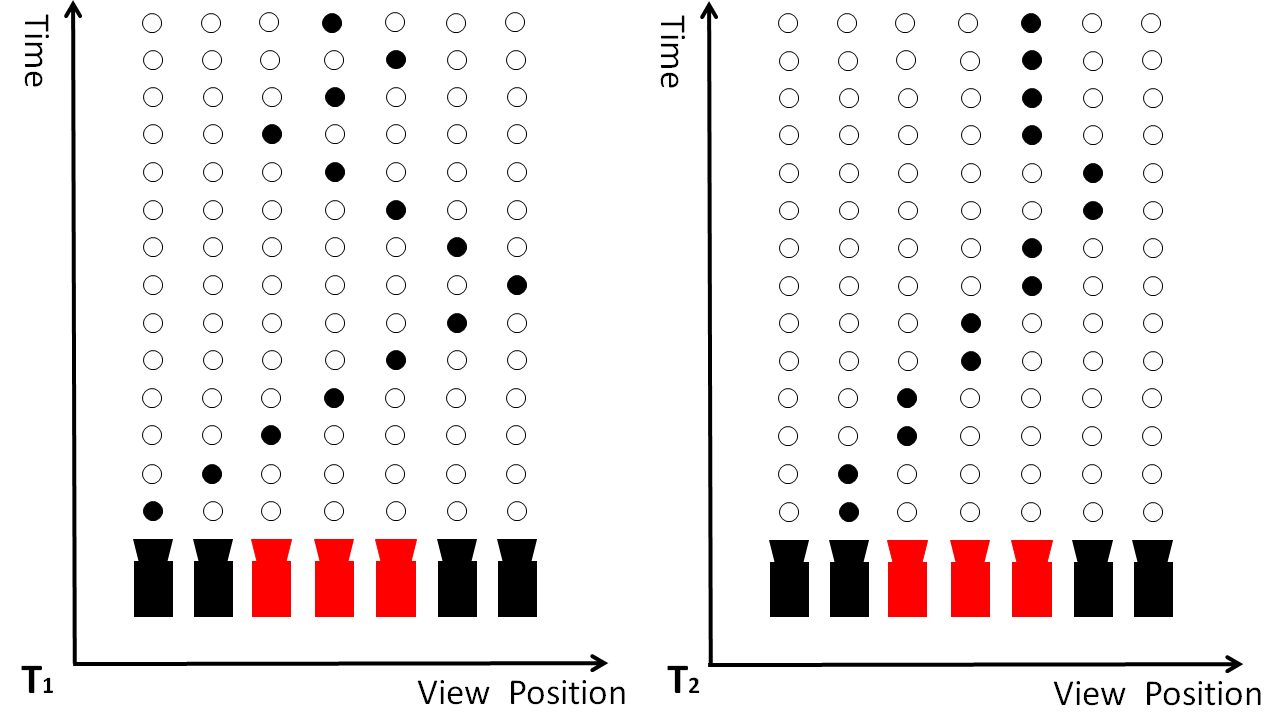}
  \caption{Description of generated trajectories. In the figure, red cameras indicate views contain important objects while the black ones represent the one mainly contain background (1) \textbf{Left $T_1$} : Sweeps (navigation path ) were constructed at a speed of one frame per view (as what is done in MPEG) (2)  \textbf{Right $T_2$} : Sweeps (navigation path )were constructed at a speed of two frames per view.}
  \label{Fig:sweep}
  \end{center}
\end{figure}

One of the purposes of this subjective experiment is to check whether semantic contents of the videos and how the navigation trajectories among views will affect the perceived quality. Therefore, different HRTs are considered in this study, generating sweeps that focus more on important objects, since human visual system tends to attach greater interest on 'Regions of Interest' (ROI) \cite{2engelke2015perceived} that contain important objects. Specifically, the following two HRTs are chosen from the pretest session considering the fact that human observers may pay more attention and even stop navigating to observe targeted objects in the video. These two HRTs are denoted with $T_1$ and $T_2$ as depicted in Fig. \ref{Fig:sweep}:  ($T_1$) An 'important-objects HRT' that first scans from the left-most to the right-most views to observe the overall contents in the video, then scans back to the views that contains the main objects and looking left and right around the central view that contain the objects several times at a velocity of one frame per view (1fpv);   ($T_2$) An 'important-objects-stay HRT' that first scans from the left-most to the right-most views to observe the overall content in the video, then scans back to the views that contain main objects at a velocity of 2fpv and finally stays in the central view that contains the main object. Due to limitation of resources, only two trajectories are considered in this study as initial exploration.



\subsection{Test Methodology}
The methodology of Absolute Category Rating with hidden reference (ACR-HR)~\cite{itu2014subjective} was adopted for the subjective experiment. Thus, the observers watched sequentially the test videos, and after each one, they provided an score using the five-level quality scale. For this, an interface with adjectives representing the whole scale was shown until the score was provided, and then, the next text video was displayed. Also, it is worth noting that each test video was shown only once and the test videos were shown to each observer in different random orders. At the beginning of the test session, an initial explanation was given to the participants indicating the purpose and how to accomplish the test. Then, a set of training videos was shown to the observers to familiarize them with the test methodology and the quality range of the content. The entire session for each observer lasts for around 30 minutes.

\subsection{Environment and Observers}
The test sequences were displayed on a professional screen TVLogic LVM401W, using a high-performance computer. Observers are provided with a tablet connected to the displayed computer for voting. The test room was set up according to the ITU recommendation BT.500~\cite{series2012methodology}, so the walls were covered by gray-color curtains and the lightning conditions were regulated accordingly to avoid annoying reflections. Also, a viewing distance of 3H (H being the height of the screen) was chosen. 

There were totally 33 participants in the subjective test, including 21 females and 12 males, with ages varying from 19 to 42 (average age of 24). Before the test, the observers were screened for correct visual acuity and color vision using the Snellen chart and Ishihara test, respectively, and all of them reported normal or corrected-to-normal vision. After the subjective test, the obtained scores were screened according to the procedure recommended by the ITU-R BT.500~\cite{series2012methodology} and the VQEG~\cite{VQEG2010b}. As a result form this screening, four observers were removed.

\section{Subjective Experiment Results and Analysis}
\label{sec:sub-ER}

\begin{figure*}[!htbp]
\subfloat[Sequence BBB Flowers with $T_1$]{\label{ex0}
\includegraphics[width=0.5\textwidth]{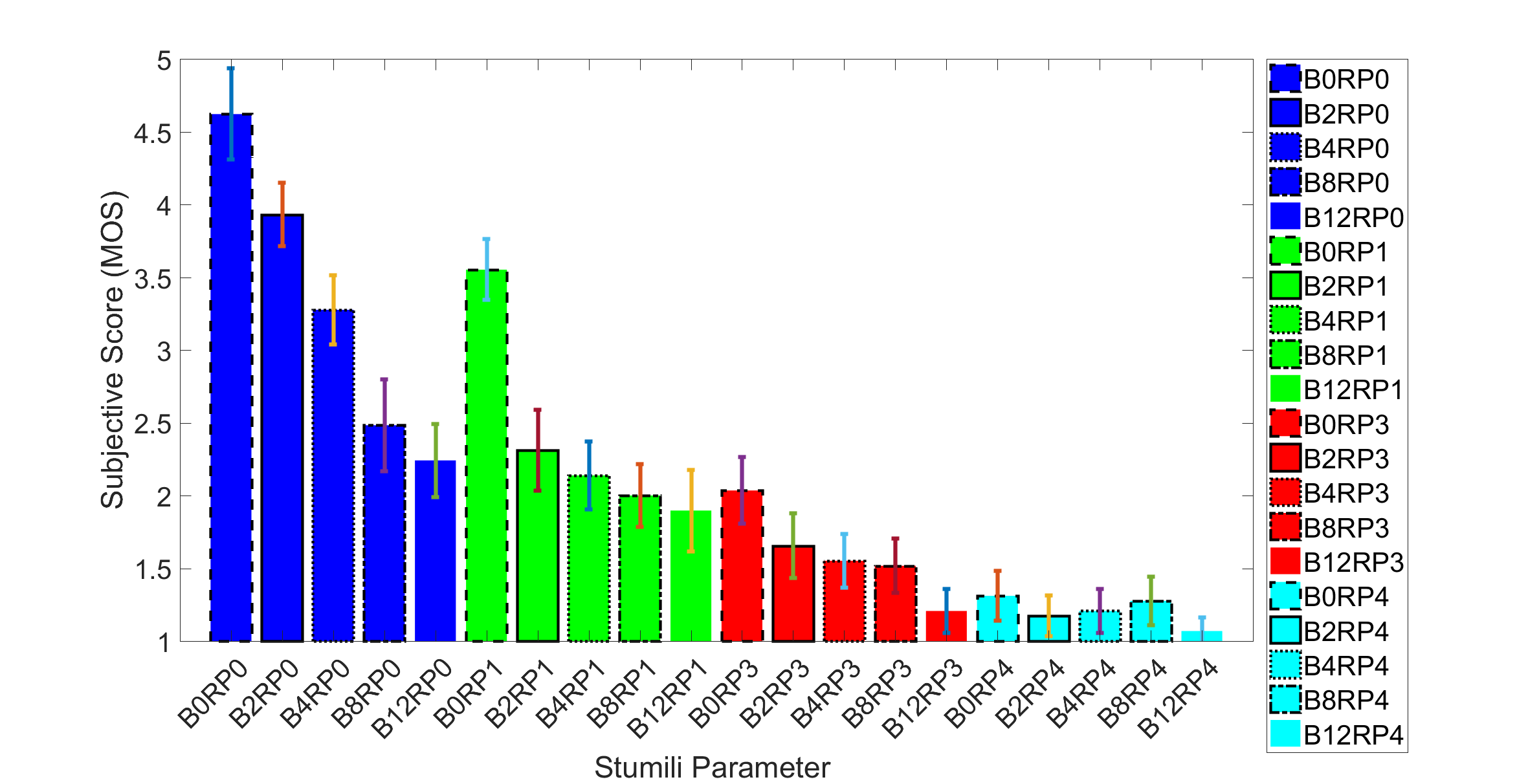}}
\subfloat[Sequence BBB Flowers with $T_2$]{\label{ex1}
\includegraphics[width=0.5\textwidth]{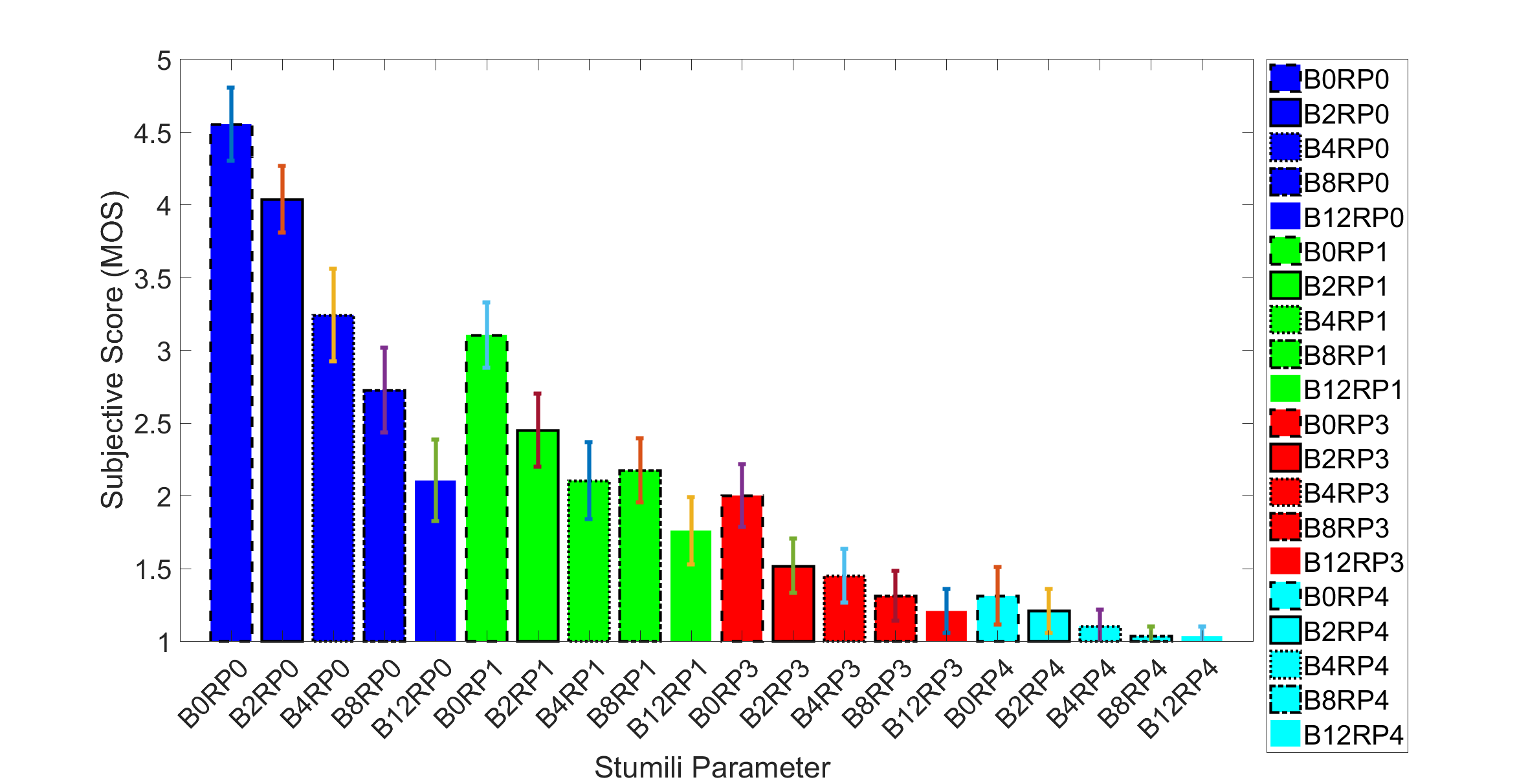}}  \\
\subfloat[Sequence Champagne with $T_1$]{\label{ex3}
\includegraphics[width=0.5\textwidth]{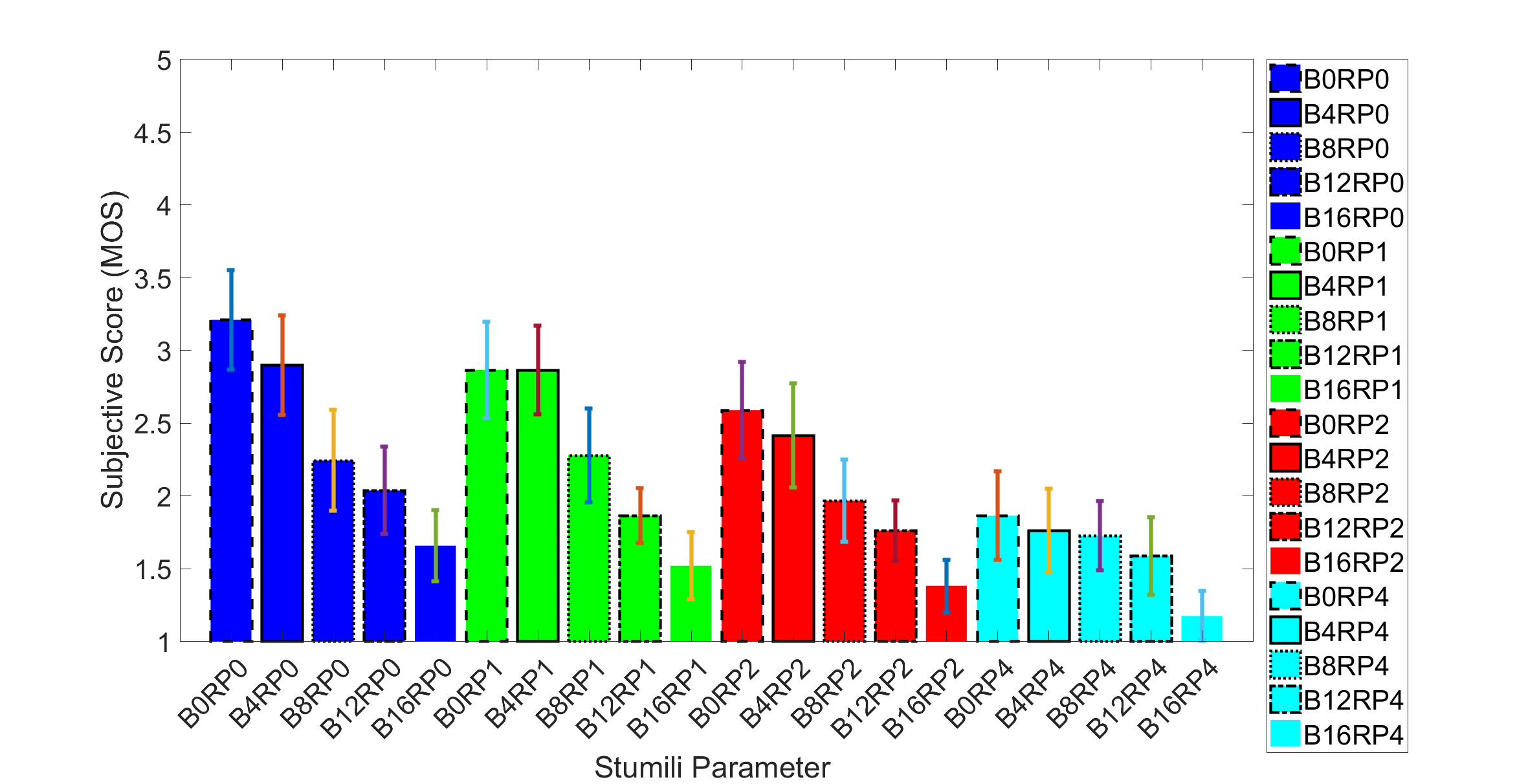}}
\subfloat[Sequence Champagne with $T_2$]{\label{ex4}
\includegraphics[width=0.5\textwidth]{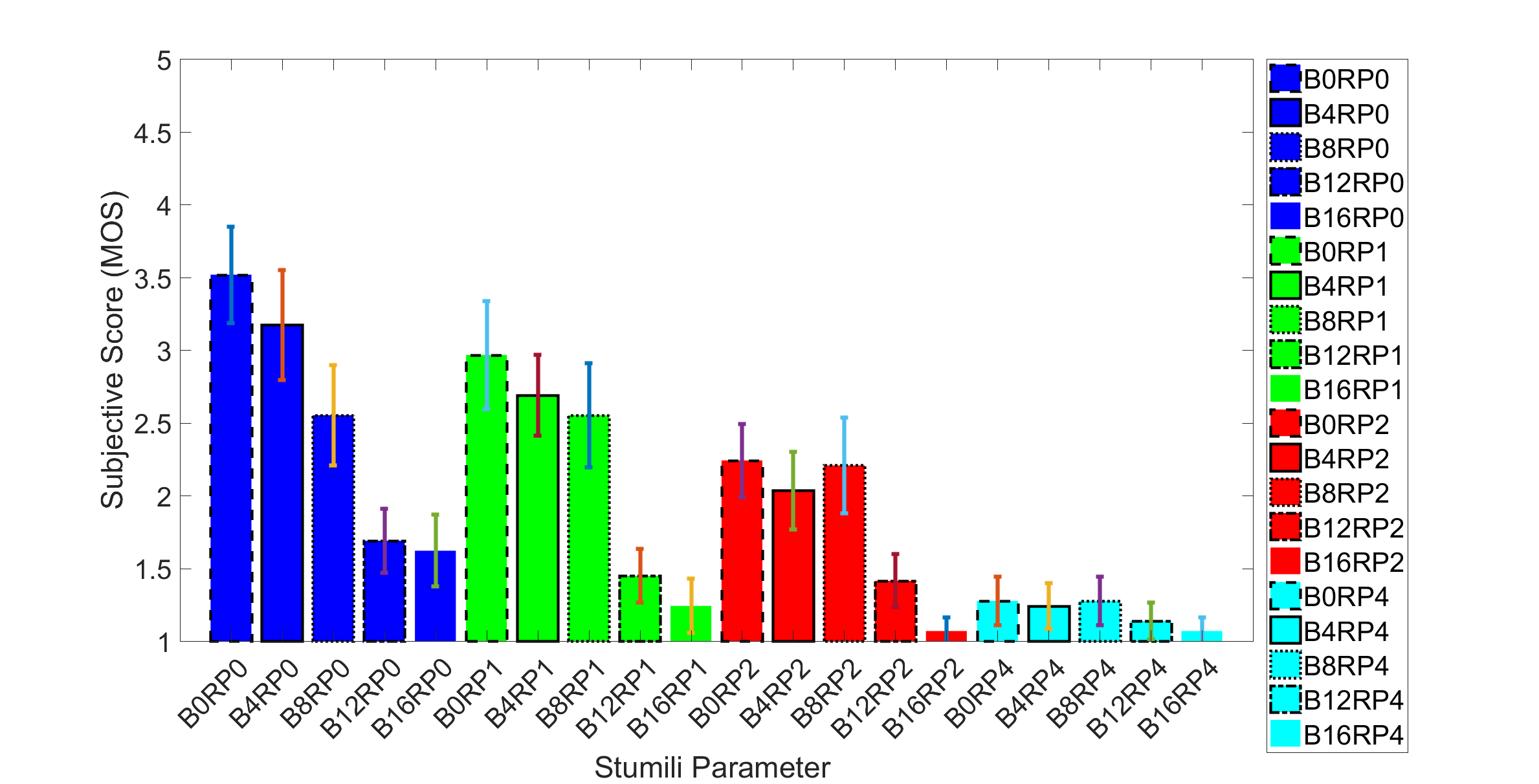}}  \\
\subfloat[Sequence Pantomime with $T_1$]{\label{ex5}
\includegraphics[width=0.5\textwidth]{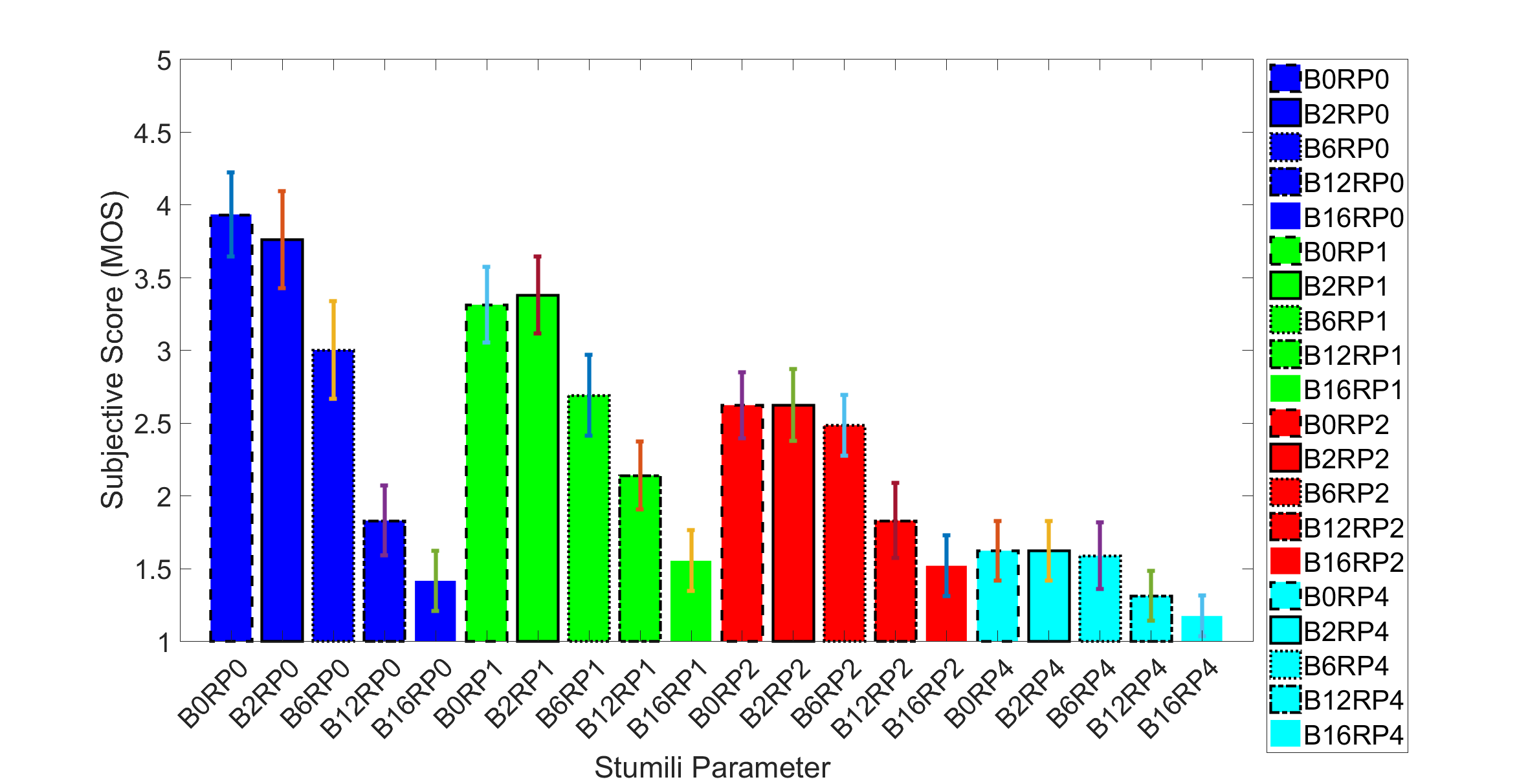}}
\subfloat[Sequence Pantomime with $T_2$]{\label{ex6}
\includegraphics[width=0.5\textwidth]{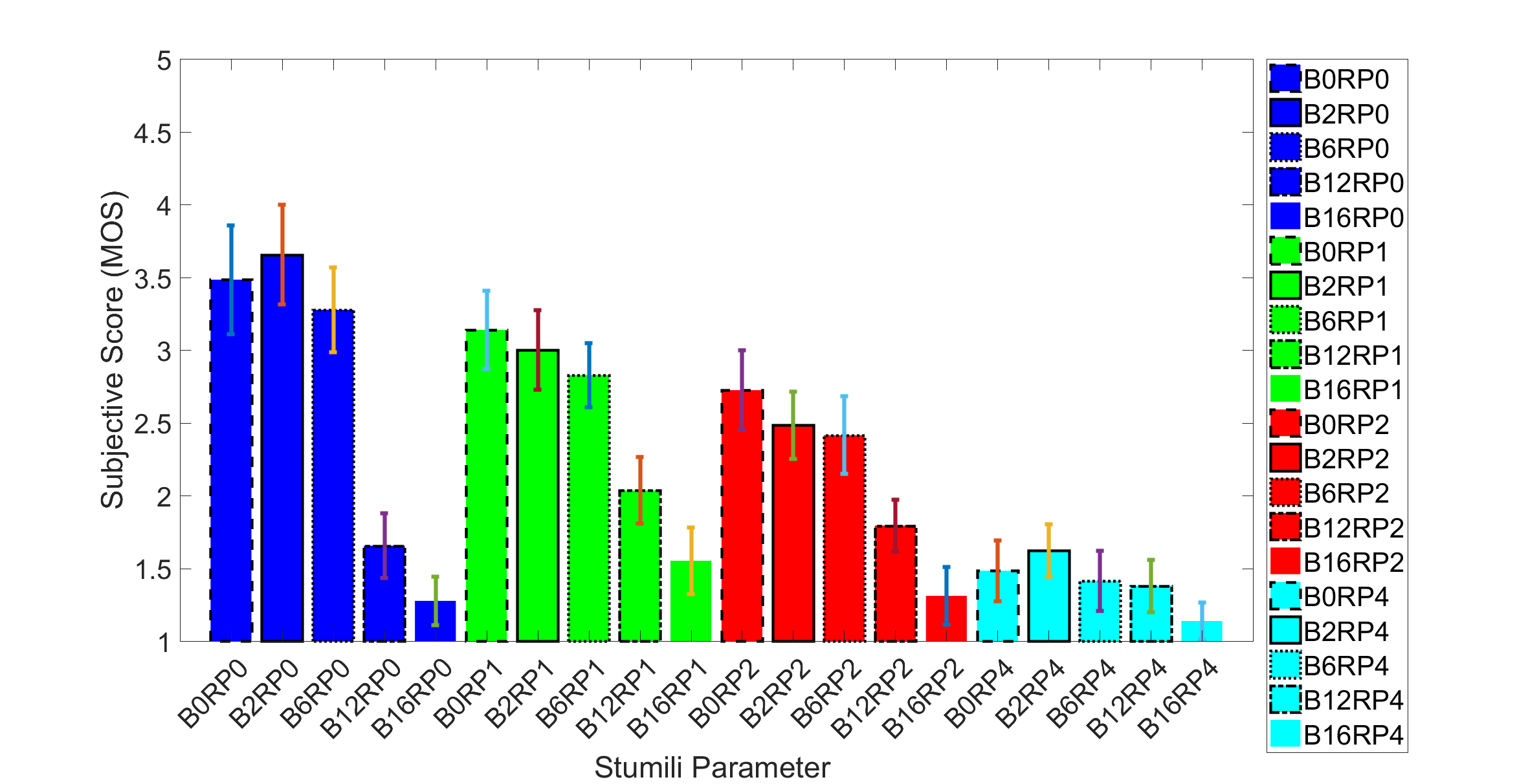}}
\caption{MOS of the sweeping sequences with different Rate-Point (RP), different baseline (B) and different sweeping trajectory (T) in the IPI-FVV Database.}
\label{fig:sub_result}
\end{figure*}

The subjective result is shown in Fig.~\ref{fig:sub_result}, where each sub-graph summarizes the mean opinion score (MOS) (with confidence intervals~\cite{series2012methodology}) for each content in each virtual sweep. Apart from MOS, the differential mean opinion score (DMOS) is also provided along with the database, computed from the hidden references according to~\cite{itu2014subjective}. As required for a quality dataset, the MOS values are well distributed covering almost the whole rating scale. In addition, in order to verify whether different Baselines (B), Rate-Points (RP) and ,specially, virtual Trajectories (T) have significant impacts on perceived quality, a three-way analysis of variance (ANOVA) was performed. From the results of this test and the results shown in Fig.~\ref{fig:sub_result}, the following main conclusions could be drawn:

\begin{itemize}
\item  At same configuration (i.e. baseline, rate-point and trajectory), the quality obtained with different contents are significantly different. 
\item The effects of view-synthesis and compression artifacts are obvious, as shown when considering how the perceived quality changes with only baseline (for a given RP), or with only bitrate (fixing the baseline). The accumulation of the effects can be also observed in the scores for the tests sequences with combined degradations. 

\item The three considered factors, specially trajectory $T$, have significant impact on the perceived quality ($p=0$ for $B$ and $RP$, and $p=0.038$ for $T$). 
\item  In terms of interaction among the considered factors, the interaction between baseline distance and coding quality has a significant effect on the MOS scores ($p=0$), as expected. 

\end{itemize}

Following are more detailed analysis of the impact of trajectory on perceived quality:
\begin{enumerate}
\item  The averaged MOS values (averaged contents 'CT', 'P', 'BBBF' and conditions) of sequences in form of $T_2$ is smaller than the one of $T_1$. Apart from ANOVA test, to further confirm the impact of trajectory on perceived quality, the database is divided into two sets (i.e., sequence with $T_1$ and with $T_2$). A t-test is conducted by taking the pairs of sequences in form of $T_1$ and $T_2$ with same baseline, rate-point configuration as input. According to the result, there is a significance difference between the quality of these two sets (i.e. $T_1$ and $T_2$). 

\item   Certain contents are more sensitive to certain trajectories. To further check whether the impact of certain trajectories depend on the content of the sequences, another t-test is conducted. More specifically, for each content, pairs of sequences that generated with the same baseline and ratepoint but different trajectory are first formed. Then, a t-test is conducted by taking the individual subjective scores (opinion scores from all the observers) of each pair of these sequences as input. According to the t-test result, for content `C', 50 \% of the pairs are of significantly different perceived quality. However, for content `CT' and `BBBF', only around 10\% of pairs are of significantly different quality. It is proven that the impact of the trajectory on quality is content dependent. In other words, `extreme trajectory' of videos with different contents are different.


\item Whether the the quality of sequence in form of one trajectory is higher than another depends also on quality range (in terms of baseline and rate-point setting).  Result of t-test taking individual subjective score of each trajectory pair as input also shows that, for content 'C' videos that in form of $T_2$ is of better quality than the ones in $T_1$ when quality is higher than a certain threshold (smaller baseline or smaller rate-point) and vise versa. For example, for content 'C' with rate-point larger than $RP_2$, sequence in form of $T_1$ is better than the one with $T_2$.

\end{enumerate}
In conclusion, it is confirmed by the subjective study that there is an impact on perceived quality from navigation trajectory. It is found that content related trajectory is able to stress the system one step further for a more extreme situation. Therefore, image/video objective metrics that is able to indicate sequences in form of one trajectory is of better quality than another is required to better push the system to its limit according to the contents. To fill out this need, a video quality metric is introduced in the next section.

\section{ Video Quality Measure for Free Viewpoint Videos} \label{sec:OVQA}
Objective quality measure that could provide more robust indication of the quality for a given HRT is required. Towards this goal, a Sketch-Token based Video Quality Measure (ST-VQM) is proposed to quantify the change of structure. 'Sketch-Token' (ST)~\cite{lim2013sketch} model  is a bag-of-words approach training a dictionary for representing the contours with contour's categories. Considering the fact that (1) content related trajectory is able to stress the system; (2) content is related to structure; (3) geometric distortions are the most annoying degradations that interrupt structure introduced by view synthesis, the main idea of the proposed method is to assess the quality of the navigation videos by quantifying to what extent the classes of contours change due to view synthesis, compression and transition among views. It is an extended version of our previous work~\cite{ling2017image} (a quality measure for image) to cope with the FVV scenario. In this version, the complex registration stage is replaced by local regions selection, and a ST-based temporal estimator is incorporated to quantify temporal artifacts.

\begin{figure*}[!htbp]
\begin{center}
 \includegraphics[width=1.5\columnwidth]{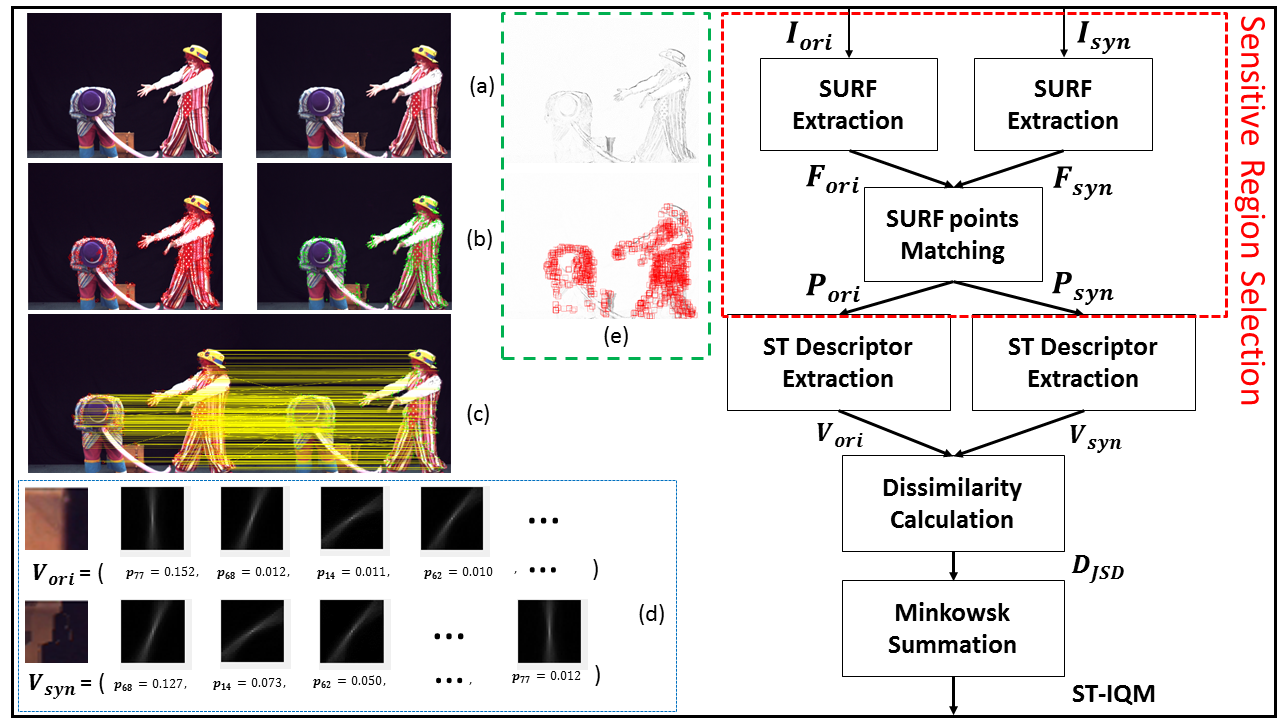}
  \caption{Overall framework of the proposed objective metric: (a) Reference image (on the left) and synthesized image (on the right); (b)Extracted SURF key-points of the reference and synthesized images; (c) Matched key-points from the reference to the synthesized image (connected with yellow lines); (d) Extracted ST feature vector of the corresponding patches and its visualization of each contour category}
  \label{Fig:overall}
  \end{center}
\end{figure*}

The improved video quality metric is consist of two parts, including a spatial metric ST-IQM as shown in Fig.~\ref{Fig:overall} and a temporal metric ST-T as shown in Fig.~\ref{Fig:temporal}. Details of each part is given in the following subsections.
 



\subsection{Sensitive Region Selection based on Interest Points Matching and Registration }
\label{ssec: LRS}

Sensitive region selection is important for the later evaluation of the quality of DIBR-based synthesized views mainly for the following reasons: (1) instead of uniform distortions distributed equally throughout the entire frame, synthesized views contains mainly local nonuniform geometric distortion; (2) distortions distributed around region of interest are less tolerant for human observers than a degradation locating at an inconspicuous area~\cite{ninassi2007does}. Meanwhile, 'poor' regions are more likely to be perceived by humans in an image with more severity than the 'good' ones. Thus images with even a small number of 'poor' regions are penalized more gravely; (3) global and local shifting of objects introduced by DIBR algorithms is a big challenge for point to point metrics like PSNR due to the mismatched correspondences. 

Interest point-based descriptors like SURF~\cite{bay2008speeded}, which reveal image's local properties and local shape information of objects are good candidates for selecting important local regions where DIBR local geometric artifacts could appear. Furthermore, later interest point matching can also be useful to compensate for consistent 'Shift of Objects' artifacts which are, to some extent, acceptable for the human visual system. 

The process of sensitive regions selection is summarized by the red dash bounding box in Fig.~\ref{Fig:overall}.
First SURF $F_{ori}$ and $F_{syn}$ points are extracted in respectively both original $I_{ori}$ and synthesized frames $I_{syn}$. Then SURF points matching between the two frames is achieved following the reference method in ~\cite{bay2008speeded} (the original frame being considered as the reference for this matching process). Pairs of interest points that have significantly different $x$ and $y$ values are discarded, being considered as not plausible matched regions from the synthesis process.
The patches $P_{ori},P_{syn}$  centered at the corresponding matched SURF points in synthesized and original images are then considered. The size of these patches is set as $35 \times 35$ to match ST formalism as introduced by \cite{lim2013sketch} (see next section). The matching relation for all patches is encoded in a matching matrix $M_{match}(x_r,y_r)=(x_m,y_m)$, where $(x_r,y_r)$ is the coordinate of the SURF point of the patch of the reference frame and $(x_m,y_m)$ is the coordinate of its matched SURF point of the patch in the synthesized frame. 

To illustrate the capability of SURF for selecting sensitive regions, one example is presented in Fig.~\ref{Fig:overall} (e). The error maps are generated with the synthesized and the reference images as introduced in~\cite{sandic2015dibr}. The darker the region the more distortions it contain, as showed in the top part of the dashed bounding green box in Fig.~\ref{Fig:overall} (e). The red bounding box represent the  sensitive regions as extracted by the proposed process. It can be observed that, as desired, the majority of regions containing severe local distortions are well identified by this process. 



\subsection{Sketch-token based Spatial Dissimilarity}
\label{sec:ST_DE}
Structures convey critical visual information and are beneficial to scene understanding, particularly the fine structures (edge) and main structures
(contour) \cite{gu2017fast,liu2012image}. Considering the process for synthesizing virtual views by DIBR methods, the key target is to transfer the occluded regions (mainly occurred at the contour of the foreground objects) in the original view to be visible in the virtual view. Measuring the variations occurred at the contours is highly related to the degradation of image quality in that use case. Consequently a method that encodes well contour would be a good candidate. The local edge-based mid-level features called 'Sketch Token' \cite{lim2013sketch} has been proposed to capture and encode contour boundaries. It is based on the idea that structure in an image patch can be described as a linear combination of "contour" patches from an universal codebook (trained once for all). 


In Lim and al. work, to train the codebook of contour patches, human subjects were asked to draw sketches as structural contours for each image in a training set. 151 classes of sketch token were formed by clustering $35 \times 35$ pixels patches from the labeled training set. 
After extracting a set of low-level features from the patches, random decision forests model was adopted to train 150 contour classifiers for the contours within patches. Each output of every trained contour classifier is the likeliness $p_i$ of the existence of one correspondence contour $i$ in the patch. The $151_{th}$ category is for patch that does not contain any structural contours, e.g. patches with only smooth texture. One can calculate $p_{151}$ with $1-\sum \limits_{i \in (1,150)} p_i$, since $\sum \limits_{i \in (1,151)} p_i=1$. Finally, the output of these 151 classifiers are concatenated to form the ST vector so that with a given pixel $(x,y)$, the corresponding patch can be represented as $V(x,y)=(p_1, p_2,...,p_{151} ) $ and the set of classifiers as the universal codebook.

In our metric, we extract the ST vectors $V_{ori}$ and $V_{syn}$ for each patches $P_{ori}$ and $P_{syn}$ of the matched SURF points pairs in matching matrix $M_{match}$.  
The dissimilarity between each matched contour vectors $V_{ori}$ and $V_{syn}$ is then computed. As the vectors contains probality with the sum of all the $p_i$ equals to 1, we propose to use Jensen--Shannon divergence as a dissimilarity measure which present the advantages to be bounded as opposed to the original Kullback--Leibler divergence.
The dissimilarity between the matched patches centering at $(x_r,y_r)$ and $(x_m,y_m)$ respectively is then calculated as 
\begin{equation}  
\begin{split}    
D_{JSD}(V_{ori},V_{syn})&= \frac{1}{2}D_{KLD}(V_{ori}(x_r,y_r),A )\\
&+\frac{1}{2}D_{KLD}(V_{syn}(x_m,y_m), A )
\end{split}
\end{equation}

Where $A=\frac{1}{2}(V_{ori}(x_r,y_r)+V_{syn}{(x_m,y_m)})$, and $D_{KLD}$ is the Kullbackâ Leibler divergence defined as 
\begin{equation} 
D_{KLD}(V_{ori},V_{syn})= \sum \limits_i V_{ori}(i)  log\frac{V_{ori}(i)}{V_{syn}(i)}
\end{equation}

In order to amplify error regions with larger dissimilarity, the Minkowski distance measure is used as pooling strategy accross sensitive regions. The spatial part of the proposed metric ST-IQM is then defined as 
\begin{equation} 
\begin{split}   
&ST\textrm{-}IQM(I_{ori},I_{syn})=\\
&\frac{ [ \sum \limits_{N}  D_{JSD}(V_{ori}(x_r,y_r),V_{syn}(x_m,y_m))^ \beta   ]^{\frac{1}{\beta}} }{N}
\end{split}  
\label{equation Minkowski}
\end{equation}
Where $N$ is the total number of matched SURF points in the frame and $\beta$ is a parameter corresponds to the $\beta-norm$ defining the $L^\beta$ vector space. 

\subsection{Sketch Token based Temporal Dissimilarity}
Sweeping between views introduces and amplifies specific temporal artifacts including flickering, temporal structure inconsistency and so on. Among them, temporal structure inconsistency is usually the most sensitive artifact for human observers since it is usually located around important moving objects and is more obvious to notice compared to other temporal artifacts.
\begin{figure}[!ht]
\begin{center}
 \includegraphics[width=1\columnwidth]{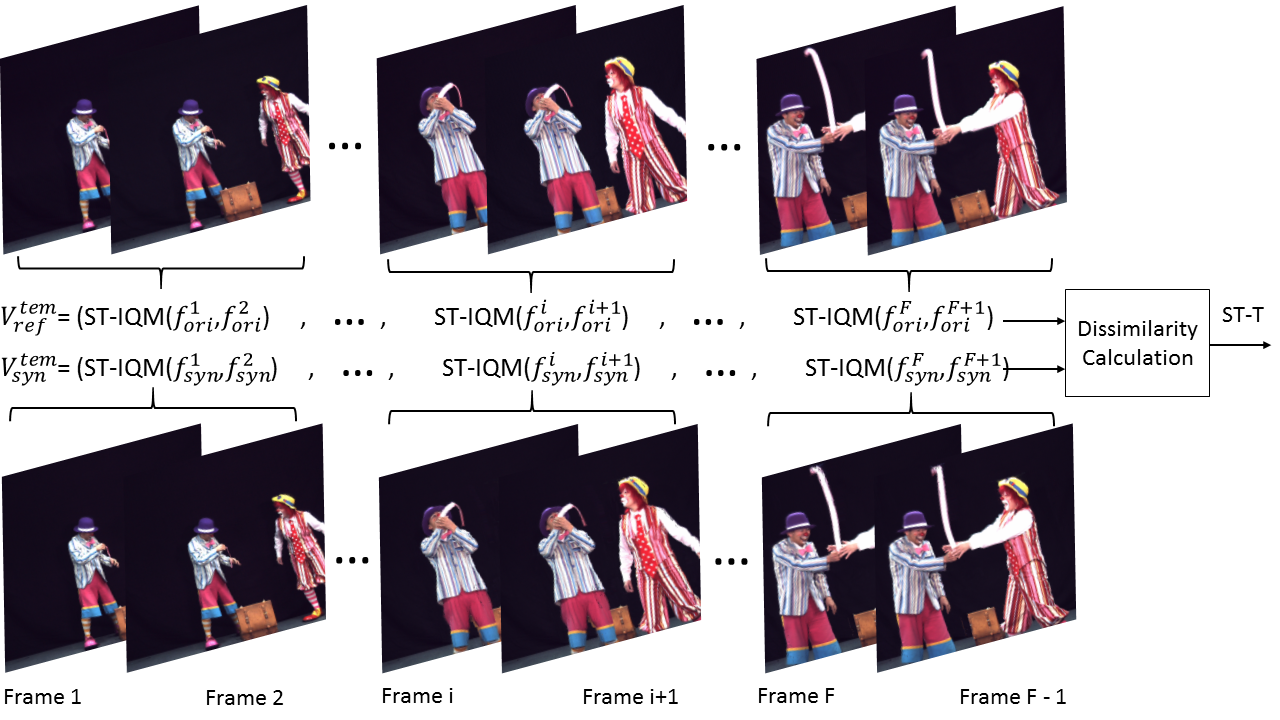}
  \caption{Diagram of Sketch Token based temporal distortion computation, where $F$ is the total frame number of the sequence. }
  \label{Fig:temporal}
  \end{center}
\end{figure}
To quantify temporal structure inconsistency, we further compute the dissimilarity score between each pair of continuous frames using the proposed Sketch-Token model introduced in section \ref{sec:ST_DE}. 
In the previous section, ST-IQM was used to quantify the difference of structure organization between two images (original purpose of this framework). It can also be used to encode and describe how structure are evolving from one frame to another along a given sequence. Temporal structure changes as observed in FVV should affect this description. This idea is exploited to refine the quality estimation in case of FVV in order to capture temporal inconsistency.

Fig. \ref{Fig:temporal} is a diagram explaining how the Sketch Token based temporal distortion is calculated. More specifically, for each pair of continuous frames of a sequence $S$, $f^i$ and $f^{i+1}$, one can compute  $ST-IQM( f^i, f^{i+1})$ using equation (\ref{equation Minkowski}). A vector $V^{tem}$ can be formed considering all frames of the sequence (each component of the vector corresponding to $ST-IQM( f^i, f^{i+1})$). We define the Sketch Token based temporal dissimilarity (ST-T) between the original and the synthesized sequences as the euclidean distance between the two temporal vectors of the original and the synthethised sequence:
\begin{equation} 
ST\textrm{-}T(S_{ori},S_{syn})= ED(V_{ori}^{tem},V_{syn}^{tem})
\end{equation}
where $ED( \cdot)$ is the euclidean distance function.
\subsection{Pooling}
With the spatial Sketch Token based score (ST-IQM) and the temporal Sketch Token based score (ST-T), it is desirable to combine them to produce an overall score. The final quality score of a synthesized sequence is defined as: 
\begin{equation} 
\begin{split}   
& ST\textrm{-}VQM= w_S \cdot ST\textrm{-}IQM   + w_T  \cdot ST\textrm{-}T + \gamma \\
\end{split} 
\label{ST-merging}
\end{equation}
where $w_S, w_T $ are two parameters used to balance the relative contributions of the spatial and temporal scores with a bias term $\gamma$. The selection and the influence of the related parameters will be given in section \ref{sec:ER}.

\section{ Experiment Results of the Proposed ST-VQM}
\label{sec:ER}
The IPI-FVV database concluded in section \ref{sec:SE} is adopted for the evaluation of the objective measures' performance. For comparison, only image/video measures designed for quality evaluation of view-synthesis artifacts are tested since commonly used metrics fail to quantify geometric distortions as already reported in \cite{sandic2016dibr,liu2015subjective,sandic2015dibr,battisti2015objective}. To compare the performances of the proposed measure with the state of the art, we firstly used the common criteria of computing Pearson correlation coefficient (PCC), Spearman's rank order correlation coefficient (SCC) and root mean squared error (RMSE) between the subjective scores and the objective ones (after applying a non-linear mapping over the measures)~\cite{VQEG2010b}. In case of image quality measures, their corresponding spatial objective scores are first calculated frame-wise, and the final object score is computed by averaging the spatial scores. 

\begin{table}[!htbp]
\centering
\caption{\label{tab:main performance}%
Performance Comparison of the Proposed measure with state-of-the-art}
\label{tab:obj_performance}
\begin{tabular}{|c|c|c|c|}\hline
 
  &\bf{PCC} &\bf{SCC} &\bf{RMSE}     \\ \hline
\multicolumn{4}{ | c |}{Image Quality Measures}  \\ \hline
3DSwIM ~\cite{battisti2015objective}  & 0,5230  & 0,5649 & 0,8640  \\
MWPSNR~\cite{sandic2015dibr}& 0,5705  & 0,8192 & 0,8304  \\
MWPSNR$r$~\cite{sandic2016dibr}   & 0,5779  & 0,8295 & 0,8252   \\
MPPSNR~\cite{sandic2015dibrMP}& 0,5706  & 0,8299 & 0,8304   \\
MPPSNR$r$~\cite{sandic2016dibr}& 0,5603  & 0,8319 & 0,8377 \\
ST-IQM  &0.8805&	0.8511&	0.4793  \\\hline
\multicolumn{4}{ | c |}{Video Quality Measures}  \\ \hline
Liu-VQM~\cite{liu2015subjective} & 0,9286  & 0,9288 & 0,3753  \\
ST-T  & 0.8336&	0.8926&	0.4837\\
ST-VQM &  \bf{0.9509} & \bf{0.9420} & \bf{0.3131}  \\
\hline
\end{tabular}
\end{table}

The overall results are summarized in Table~\ref{tab:obj_performance} and the best performance values are marked in bold. 
As it can be observed from Table~\ref{tab:obj_performance}, ST-VQM, Liu-VQM are the two best performing metrics, with PCC equals to 0.9509, 0.9286 correspondingly. To analyze if the differences between those values are significant, a T-test was carried out taking the difference of the predicted score between DMOS and Liu-VQM, and the one between DMOS and ST-VQM as inputs. The results showed that  our proposed metric significantly outperform the second best performing Liu-VQM. As it can be observed, the performance of the image metrics, including MW-PSNR and MP-PSNR, is very limited, which can be due to (1) they over-penalize the consistent shifting artifacts, and (2) these measures do not take temporal distortions into account. 

As it has been verified in the subjective experimental results, navigation scan-paths affect the perceived quality. Therefore, it is important for an objective metric to point out whether the perceived quality using a given trajectory is better than using other trajectories. As thus, the metric can be used to evaluate the limit of the system in worse navigation situations. To this end, the Krasula performance criteria \cite{krasula2016accuracy,hanhart2016benchmark} is used to assess the ability of objective measures to estimate whether one trajectory is better than another with the same rate-point and baseline configurations in terms of perceived quality. Pairs of sequences generated with the same configurations but in form of $T_1$ and $T_2$ in the dataset are selected to calculate the area under the ROC curve of the 'Better vs. Worse' categories (AUC-BW), area under the ROC curve of the 'Different vs. Similar' category (AUC-DS), and percentage of correct classification (CC) (see~\cite{krasula2016accuracy,hanhart2016benchmark} for more details). More specifically, since pairs are collected in form of $(T_1,T_2 )$ with other parameters fixed, if one metric obtain higher AUC-BW, it shows more capability to indicate that sequences with certain trajectory are better/worse than with another. Similarly, if the metric obtain higher AUC-DS, then it can better tell whether the quality of sequences in form of one trajectory is different/similar to the ones in form of another trajectory. Results are reported in Table~\ref{tab:ST_VQA_lukas}. As it can be observed, the proposed metric obtain the best performance in terms of the three evaluation measures. It is proven that the proposed ST-VQM is able to quantify temporal artifacts introduced by views switch. More importantly, ST-VQM is the most promising metric in telling sequence generated in form of which trajectory is of better quality than the others.

\begin{table}[]
\centering
\caption{Performance comparison of metrics for distinguishing sequence in different trajectories}
\label{tab:ST_VQA_lukas}
\begin{tabular}{|c|c|c|c|}\hline
    & \bf{AUC-DS}  & \bf{AUC-BW} & \bf{CC}          \\ \hline
\multicolumn{4}{ | c |}{Image Quality Metrics}  \\ \hline
3DSwIM ~\cite{battisti2015objective}  & 0.4603 & 0.8311 & \bf{0.8667} \\
MWPSNR~\cite{sandic2015dibr}   & 0.5571 & 0.6889 & 0.6000 \\
MWPSNR$r$\cite{sandic2016dibr} & 0.5317 & 0.6933 & 0.6667 \\
MPPSNR~\cite{sandic2015dibrMP} & 0.5079 & 0.7022 & 0.6667 \\
MPPSNR$r$\cite{sandic2016dibr}& 0.5238 & 0.6933 & 0.6667 \\

ST-IQM   & 0.5016 & 0.7244 & 0.6000 \\ \hline
\multicolumn{4}{ | c |}{Video Quality Metrics}  \\ \hline
Liu-VQM ~\cite{liu2015subjective}      & 0.6270 & 0.8311 & 0.7333 \\
ST-T             & 0.5857 & 0.8800 & 0.8000 \\
ST-VQM           & \bf{0.6762} & \bf{0.8933} & \bf{0.8667} \\\hline
\end{tabular}
\end{table}

\subsection{Selection of Parameters}
\begin{figure}[!htbp]

\label{fig:para_pool_STVQA}
\subfloat[ ]{
\includegraphics[width=0.16\textwidth]{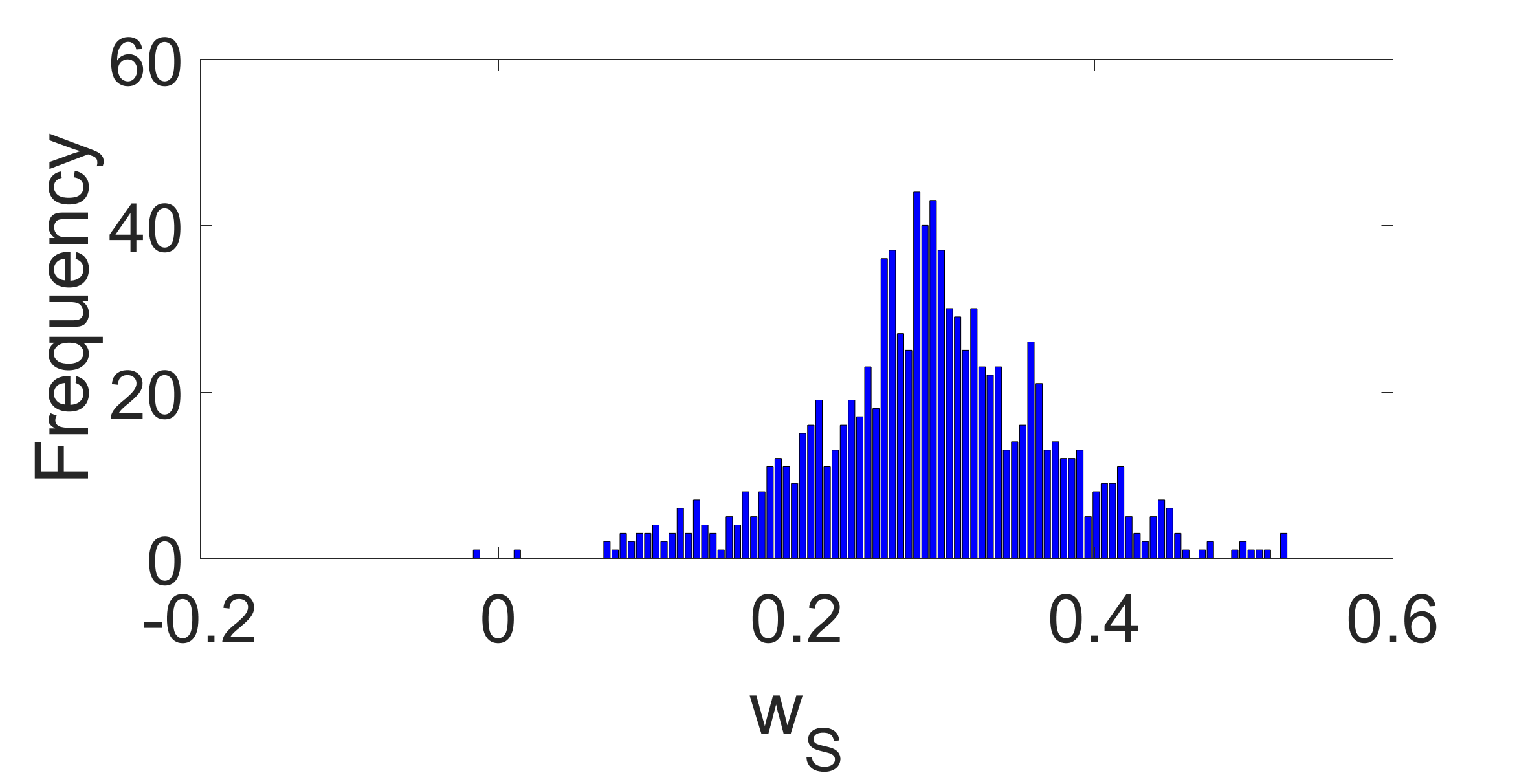}}
\subfloat[ ]{
\includegraphics[width=0.16\textwidth]{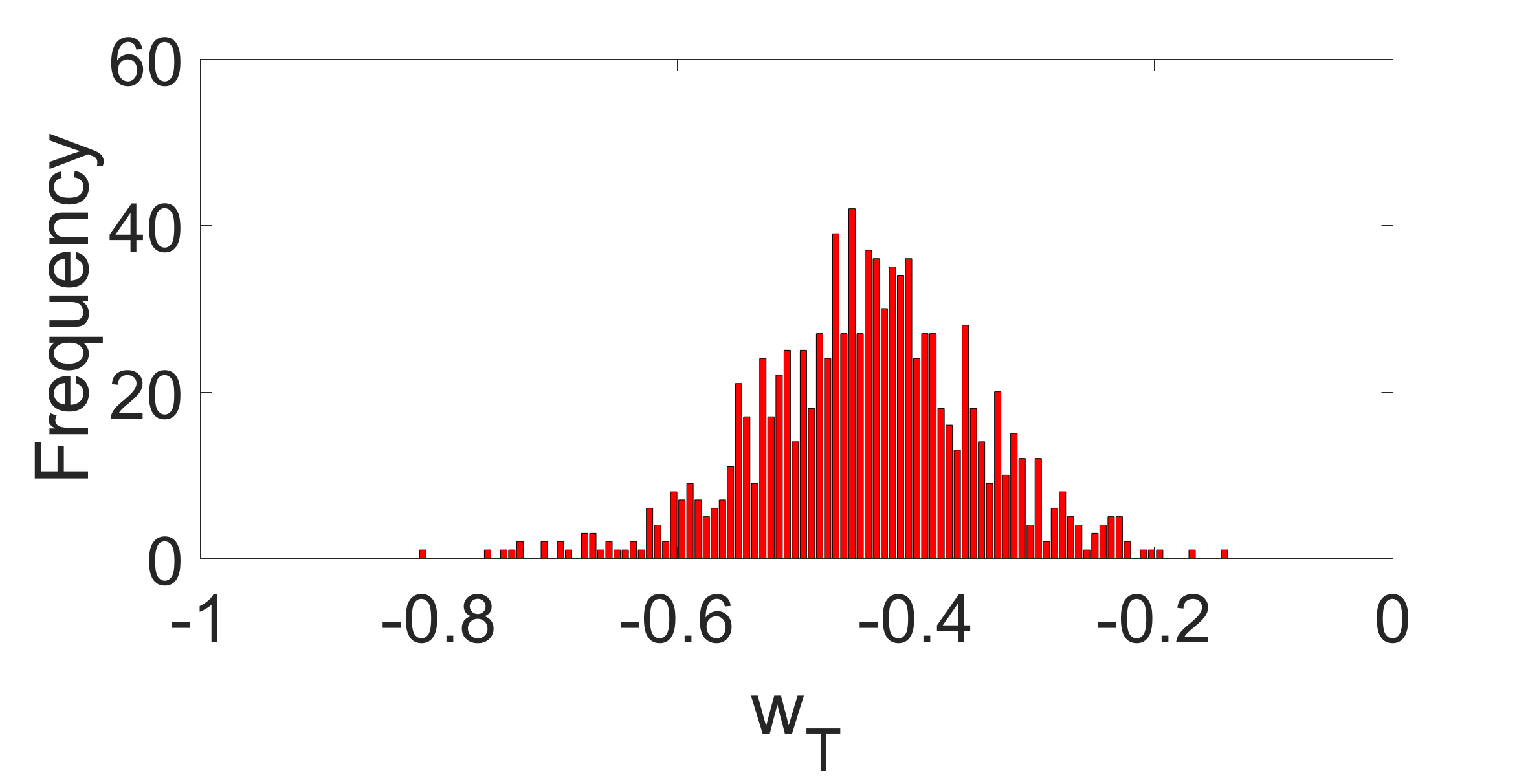}}
\subfloat[ ]{
\includegraphics[width=0.16\textwidth]{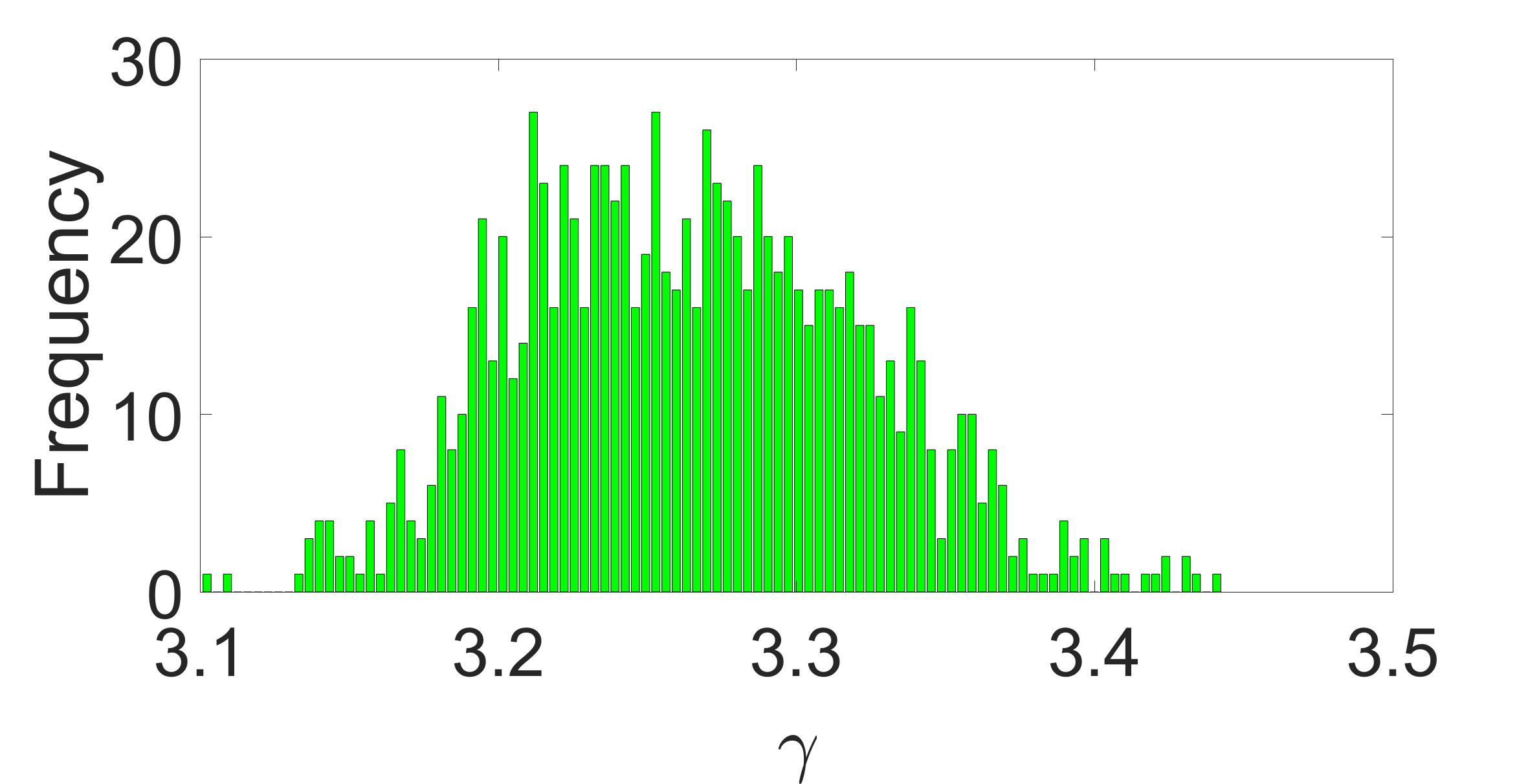}}
 \\
\subfloat[]{
\includegraphics[width=0.5\textwidth]{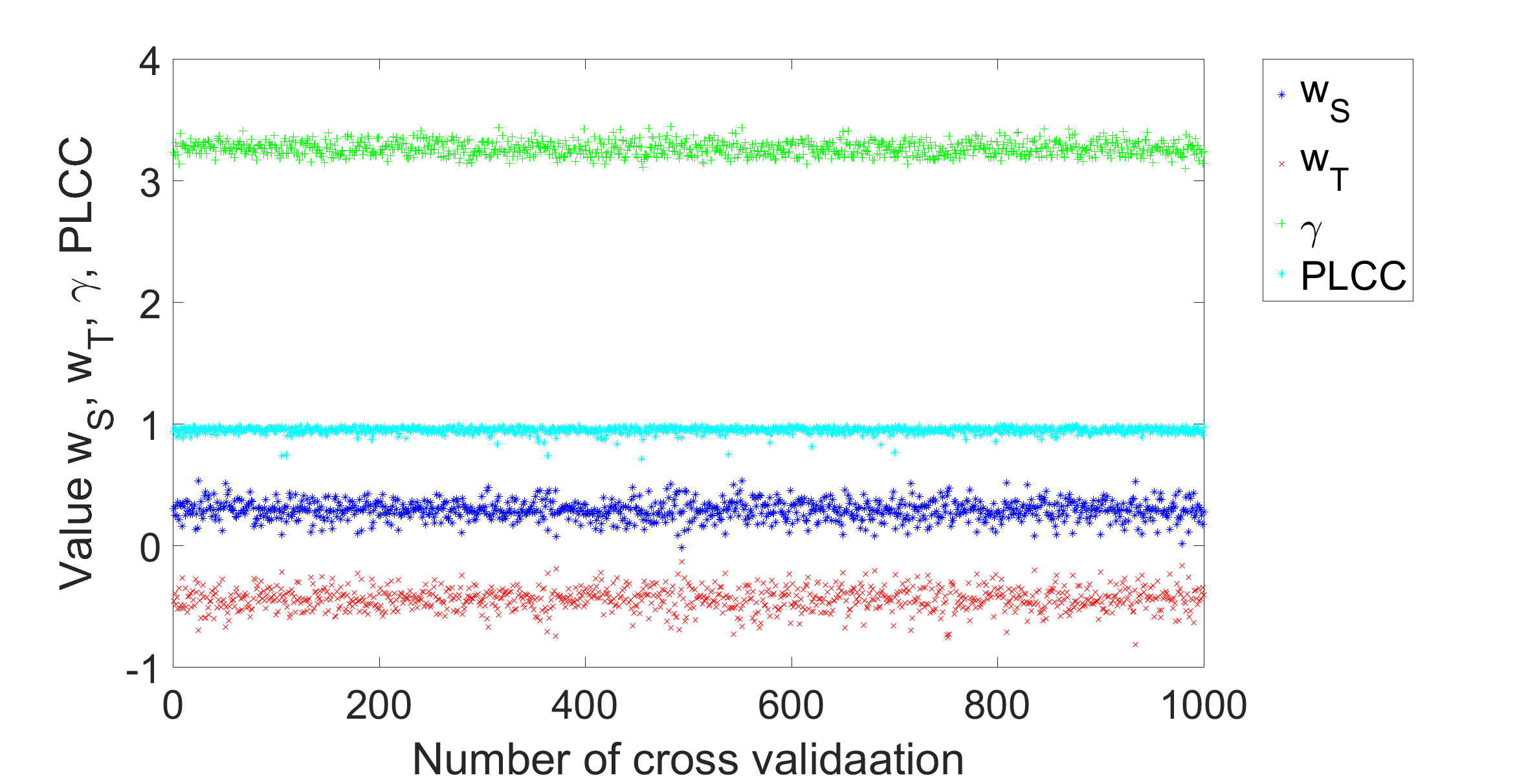}}
\\
 
\caption{ (d) Values of $w_S, w_T, \gamma$ and their corresponding PLCC across 1000 times cross validation.}
\end{figure}

It would be desirable that the performance of a VQM  does not vary significantly with a slight change of the parameters. In this section, an analysis on the selection of the parameter of the proposed metric is presented. In order to properly select $w_S, w_T$ and $\gamma$ in equation~(\ref{ST-merging}), as well as to check the performance dependency of the parameters, a 1000 times cross validation is conducted. More specifically, the entire database is separated into a training set (80\%) and testing set (20\%) 1000 times, and the most frequently occurred value will be selected for the corresponding parameter. Before the validation test, we first multiply $ST\textrm{-}IQM$ by $10^{10}$ and $ST\textrm{-}T$ by $10^5$ so that the difference between the corresponding parameter $w_S, w_T$ will be smaller making easier for latter visualization (it has to be pointed out that this operation does not change the performance).  The values of the three parameters with the corresponding PCC value across of 1000 times cross validation are shown in Fig.~\ref{fig:para_pool_STVQA} (d). It can be observed that both the values of the three parameters and the performance do not change significantly throughout 1000 times, which verifies the fact that the performance of the metric does not change dramatically along with the modification of the parameters. 
Fig.~\ref{fig:para_pool_STVQA} (a)-(b) depicts the histograms of frequencies of the three parameters' values relatively. As it can be seen that $w_S = 0.28, w_T= -0.43 $ and $\gamma= 3.26$ are the three most frequent value among 1000 times. They are thus selected and fixed for reporting the final performance in Table \ref{tab:main performance} and \ref{tab:ST_VQA_lukas}. The mean value of PCC, SROCC, and RMSE of the proposed metric across the 1000 times is  0.9513, 0.9264 and 0.2895 correspondingly, which are close to the performance values reported in Table~\ref{tab:main performance} with the selected configuration.

Subsequently, the performance dependency of the proposed algorithm on the exponent variable $\beta$ in equation (\ref{equation Minkowski}) and the distance approaches has been reported and examined in \cite{ling2017image}. Therefore, in this paper, the same $\beta = 4$ and the Jensen Shannon divergence are selected.

\section{Conclusion }
\label{sec:con}
In this paper, aiming at better quantifying the specific distortions in sequences generated for FVV systems, both subjective and objective analyses have been conducted. On one side, in the subjective study, different configurations of compression and view-synthesis have been considered, which are the two main sources of degradations in FVV. In addition, following the approach of using simulating navigation trajectories that the users of immersive media may employ to  explore the content, two different trajectories  (referred as Hypothetical Rendering Trajectories) have been used to study their impact on the perceived quality. Knowing these posible effects, may help on the identification of critical trajectories that may be more suitable to carry out quality evaluation studies related to the benchmark of systems in the worst cases, Also, it must be pointed out that the sweeps that generated in this test focus more on views that contain region of interest (e.g. moving objects) in videos since human observers are more interested in them and even stop navigating after these regions show up. By analyzing the subjective results, we find that the way of how the trajectories are generated does affect the perceived quality. In addition, the dataset generated for the subjective tests (called IPI-FVV), along worth the obtained scores is made available for the research community in the field. On the other side, in the objective study, a Sketch-Token-based VQA metric is proposed by checking how the classes of contours change between the reference and the degraded sequences spatially and temporally. The results of the experiments conducted on IPI-FVV database has shown that the performance of proposed ST-VQM is promising. More importantly, ST-VQM is the best performing metric in predicting if sequences based on a given trajectory are of higher/lower quality than sequences based on other trajectories, with respect to subjective scores. Finally, in the future, (1) related subjective and objective studies will be extended for light-field applications; (2) ST-VQM will be improved as no reference metric.

\bibliographystyle{IEEEtran}
\bibliography{bare_jrnl}

\begin{thebibliography}{10}
\providecommand{\url}[1]{#1}
\csname url@samestyle\endcsname
\providecommand{\newblock}{\relax}
\providecommand{\bibinfo}[2]{#2}
\providecommand{\BIBentrySTDinterwordspacing}{\spaceskip=0pt\relax}
\providecommand{\BIBentryALTinterwordstretchfactor}{4}
\providecommand{\BIBentryALTinterwordspacing}{\spaceskip=\fontdimen2\font plus
\BIBentryALTinterwordstretchfactor\fontdimen3\font minus
  \fontdimen4\font\relax}
\providecommand{\BIBforeignlanguage}[2]{{%
\expandafter\ifx\csname l@#1\endcsname\relax
\typeout{** WARNING: IEEEtran.bst: No hyphenation pattern has been}%
\typeout{** loaded for the language `#1'. Using the pattern for}%
\typeout{** the default language instead.}%
\else
\language=\csname l@#1\endcsname
\fi
#2}}
\providecommand{\BIBdecl}{\relax}
\BIBdecl

\bibitem{tanimoto2012ftv}
M.~Tanimoto, ``Ftv: Free-viewpoint television,'' \emph{Signal Processing: Image
  Communication}, vol.~27, no.~6, pp. 555--570, 2012.

\bibitem{muller20113}
K.~Muller, P.~Merkle, and T.~Wiegand, ``3-d video representation using depth
  maps,'' \emph{Proceedings of the IEEE}, vol.~99, no.~4, pp. 643--656, 2011.

\bibitem{fehn2006interactive}
C.~Fehn, R.~De~La~Barre, and S.~Pastoor, ``Interactive 3-dtv-concepts and key
  technologies,'' \emph{Proceedings of the IEEE}, vol.~94, no.~3, pp. 524--538,
  2006.

\bibitem{fehn2004depth}
C.~Fehn, ``Depth-image-based rendering (dibr), compression, and transmission
  for a new approach on 3d-tv,'' in \emph{Electronic Imaging 2004}.\hskip 1em
  plus 0.5em minus 0.4em\relax International Society for Optics and Photonics,
  2004, pp. 93--104.

\bibitem{wang2010asymmetric}
L.-H. Wang, X.-J. Huang, M.~Xi, D.-X. Li, and M.~Zhang, ``An asymmetric edge
  adaptive filter for depth generation and hole filling in 3dtv,'' \emph{IEEE
  Transactions on Broadcasting}, vol.~56, no.~3, pp. 425--431, 2010.

\bibitem{zhang2014efficient}
Y.~Zhang, S.~Kwong, S.~Hu, and C.-C.~J. Kuo, ``Efficient multiview depth coding
  optimization based on allowable depth distortion in view synthesis,''
  \emph{IEEE Transactions on Image Processing}, vol.~23, no.~11, pp.
  4879--4892, 2014.

\bibitem{chung2014bit}
T.-Y. Chung, J.-Y. Sim, and C.-S. Kim, ``Bit allocation algorithm with novel
  view synthesis distortion model for multiview video plus depth coding,''
  \emph{IEEE Transactions on Image Processing}, vol.~23, no.~8, pp. 3254--3267,
  2014.

\bibitem{carballeira2015subjective}
P.~Carballeira, J.~Guti{\'e}rrez, F.~Mor{\'a}n, J.~Cabrera, and N.~Garc{\'\i}a,
  ``Subjective evaluation of super multiview video in consumer 3d displays,''
  in \emph{International Workshop on Quality of Multimedia Experience (QoMEX)},
  Costa Navarino, Greece, May 2015, pp. 1--6.

\bibitem{domanski2016optimization}
M.~Doma{\'n}ski, A.~Dziembowski, A.~Grzelka, and D.~Mieloch, ``Optimization of
  camera positions for free-navigation applications,'' in \emph{Signals and
  Electronic Systems (ICSES), 2016 International Conference on}.\hskip 1em plus
  0.5em minus 0.4em\relax IEEE, 2016, pp. 118--123.

\bibitem{hanhart2014free}
P.~Hanhart, E.~Bosc, P.~Le~Callet, and T.~Ebrahimi, ``Free-viewpoint video
  sequences: A new challenge for objective quality metrics,'' in \emph{IEEE
  International Workshop on Multimedia Signal Processing (MMSP)}, Jakarta,
  Indonesia, Sep. 2014, pp. 1--6.

\bibitem{merkle2009effects}
P.~Merkle, Y.~Morvan, A.~Smolic, D.~Farin, K.~Mueller, P.~de~With, and
  T.~Wiegand, ``The effects of multiview depth video compression on multiview
  rendering,'' \emph{Signal Processing: Image Communication}, vol.~24, no.~1,
  pp. 73--88, 2009.

\bibitem{kilner2009objective}
J.~Kilner, J.~Starck, J.-Y. Guillemaut, and A.~Hilton, ``Objective quality
  assessment in free-viewpoint video production,'' \emph{Signal Processing:
  Image Communication}, vol.~24, no.~1, pp. 3--16, 2009.

\bibitem{oh2010virtual}
K.-J. Oh, S.~Yea, A.~Vetro, and Y.-S. Ho, ``Virtual view synthesis method and
  self-evaluation metrics for free viewpoint television and 3d video,''
  \emph{International Journal of Imaging Systems and Technology}, vol.~20,
  no.~4, pp. 378--390, 2010.

\bibitem{do2009quality}
L.~Do, S.~Zinger, Y.~Morvan, and P.~H. de~With, ``Quality improving techniques
  in dibr for free-viewpoint video,'' in \emph{3DTV Conference: The True
  Vision-Capture, Transmission and Display of 3D Video, 2009}.\hskip 1em plus
  0.5em minus 0.4em\relax IEEE, 2009, pp. 1--4.

\bibitem{liu2009compression}
Y.~Liu, S.~Ma, Q.~Huang, D.~Zhao, W.~Gao, and N.~Zhang, ``Compression-induced
  rendering distortion analysis for texture/depth rate allocation in 3d video
  compression,'' in \emph{Data Compression Conference, 2009. DCC'09.}\hskip 1em
  plus 0.5em minus 0.4em\relax IEEE, 2009, pp. 352--361.

\bibitem{liu2015subjective}
X.~Liu, Y.~Zhang, S.~Hu, S.~Kwong, C.-C.~J. Kuo, and Q.~Peng, ``Subjective and
  objective video quality assessment of 3d synthesized views with texture/depth
  compression distortion,'' \emph{IEEE Transactions on Image Processing},
  vol.~24, no.~12, pp. 4847--4861, 2015.

\bibitem{Battisti2016}
F.~Battisti and P.~{Le Callet}, ``{Quality Assessment in the context of FTV:
  challenges, first answers and open issues},'' \emph{IEEE COMSOC MMTC
  Communications - Frontiers}, vol.~11, no.~2, pp. 22--26, Mar. 2016.

\bibitem{Dricot2015}
A.~Dricot, J.~Jung, M.~Cagnazzo, B.~Pesquet, F.~Dufaux, P.~T. Kov{\'{a}}cs, and
  V.~K. Adhikarla, ``{Subjective evaluation of Super Multi-View compressed
  contents on high-end light-field 3D displays},'' \emph{Signal Processing:
  Image Communication}, vol.~39, pp. 369--385, Nov. 2015.

\bibitem{Revised_summary2016}
O.~Stankiewicz, K.~Wegner, T.~Senoh, G.~Lafruit, V.~Baroncini, and M.~Tanimoto,
  ``Revised summary of call for evidence on free-viewpoint television:
  Super-multiview and free navigation,'' \emph{ISO/IEC JTC1/SC29/WG11
  MPEG2016/N16523}, Oct. 2016.

\bibitem{carballeira2017multiview}
P.~Carballeira, J.~Guti{\'e}rrez, F.~Mor{\'a}n, J.~Cabrera, F.~Jaureguizar, and
  N.~Garc{\'\i}a, ``Multiview perceptual disparity model for super multiview
  video,'' \emph{IEEE Journal of Selected Topics in Signal Processing},
  vol.~11, no.~1, pp. 113--124, 2017.

\bibitem{Recio2017}
R.~Recio, P.~Carballeira, J.~Gutierrez, and N.~Garcia, ``{Subjective Assessment
  of Super Multiview Video with Coding Artifacts},'' \emph{IEEE Signal
  Processing Letters}, vol.~24, no.~6, pp. 868--871, Jun. 2017.

\bibitem{Hinds2017}
A.~T. Hinds, D.~Doyen, and P.~Carballeira, ``{Toward the realization of six
  degrees-of-freedom with compressed light fields},'' in \emph{2017 IEEE
  International Conference on Multimedia and Expo (ICME)}, Hong Kong, China,
  Jul. 2017, pp. 1171--1176.

\bibitem{bosc2011towards}
E.~Bosc, R.~Pepion, P.~Le~Callet, M.~Koppel, P.~Ndjiki-Nya, M.~Pressigout, and
  L.~Morin, ``Towards a new quality metric for 3-d synthesized view
  assessment,'' \emph{IEEE Journal of Selected Topics in Signal Processing},
  vol.~5, no.~7, pp. 1332--1343, Nov. 2011.

\bibitem{Bosc2013b}
E.~Bosc, P.~{Le Callet}, L.~Morin, and M.~Pressigout, ``{Visual Quality
  Assessment of Synthesized Views in the Context of 3D-TV},'' in \emph{3D-TV
  System with Depth-Image-Based Rendering}.\hskip 1em plus 0.5em minus
  0.4em\relax New York, NY: Springer New York, 2013, pp. 439--473.

\bibitem{bosc2013quality}
E.~Bosc, P.~Hanhart, P.~Le~Callet, and T.~Ebrahimi, ``A quality assessment
  protocol for free-viewpoint video sequences synthesized from decompressed
  depth data,'' in \emph{International Workshop on Quality of Multimedia
  Experience (QoMEX)}, Klagenfurt, Germany, Jul. 2013, pp. 100--105.


\bibitem{krasula2016accuracy}
L.~Krasula, K.~Fliegel, P.~Le~Callet, and M.~Klima, ``On the accuracy 
   of objective image and video quality models: New methodology for performance evaluation,'' in \emph{International Workshop on Quality of Multimedia
  Experience (QoMEX)}, 2016.



\bibitem{hanhart2016benchmark}
H.~Philippe, L.~Krasula, P.~Le~Callet, and T.~Ebrahimi, ``How to benchmark objective quality metrics from paired comparison data?,'' in \emph{International Workshop on Quality of Multimedia
  Experience (QoMEX)}, 2016.


\bibitem{sandic2016dibr}
D.~Sandic-Stankovic, D.~Kukolj, and P.~Le~Callet, ``DIBR-synthesized image quality assessment based on morphological multi-scale approach,'' \emph{EURASIP Journal on Image and Video Processing}, 2016.


\bibitem{SSIM}
Z.~Wang, A.~C. Bovik, H.~R. Sheikh, and E.~P. Simoncelli, ``Image quality
  assessment: from error visibility to structural similarity,'' \emph{IEEE
  transactions on image processing}, vol.~13, no.~4, pp. 600--612, 2004.

\bibitem{conze2012objective}
P.-H. Conze, P.~Robert, and L.~Morin, ``Objective view synthesis quality
  assessment,'' in \emph{IS\&T/SPIE Electronic Imaging}.\hskip 1em plus 0.5em
  minus 0.4em\relax International Society for Optics and Photonics, 2012, pp.
  82\,881M--82\,881M.

\bibitem{battisti2015objective}
F.~Battisti, E.~Bosc, M.~Carli, P.~Le~Callet, and S.~Perugia, ``Objective image
  quality assessment of 3d synthesized views,'' \emph{Signal Processing: Image
  Communication}, vol.~30, pp. 78--88, 2015.

\bibitem{sandic2015dibr}
D.~Sandi{\'c}-Stankovi{\'c}, D.~Kukolj, and P.~Le~Callet, ``Dibr synthesized
  image quality assessment based on morphological wavelets,'' in \emph{Quality
  of Multimedia Experience (QoMEX), 2015 Seventh International Workshop
  on}.\hskip 1em plus 0.5em minus 0.4em\relax IEEE, 2015, pp. 1--6.

\bibitem{sandic2015dibrMP}
D.~Sandic-Stankovic, D.~Kukolj, and P.~Le~Callet, ``Dibr synthesized image
  quality assessment based on morphological pyramids,'' in \emph{2015
  3DTV-Conference: The True Vision-Capture, Transmission and Display of 3D
  Video (3DTV-CON)}.\hskip 1em plus 0.5em minus 0.4em\relax IEEE, 2015, pp.
  1--4.

\bibitem{tsai2013quality}
C.-T. Tsai and H.-M. Hang, ``Quality assessment of 3d synthesized views with
  depth map distortion,'' in \emph{Visual Communications and Image Processing
  (VCIP), 2013}.\hskip 1em plus 0.5em minus 0.4em\relax IEEE, 2013, pp. 1--6.

\bibitem{zhao2010perceptual}
Y.~Zhao and L.~Yu, ``A perceptual metric for evaluating quality of synthesized
  sequences in 3dv system,'' in \emph{Proc. SPIE}, vol. 7744, 2010, p. 77440X.

\bibitem{ekmekcioglu2010depth}
E.~Ekmekcioglu, S.~Worrall, D.~De~Silva, A.~Fernando, and A.~M. Kondoz, ``Depth
  based perceptual quality assessment for synthesized camera viewpoints,'' in
  \emph{International Conference on User Centric Media}.\hskip 1em plus 0.5em
  minus 0.4em\relax Springer, 2010, pp. 76--83.

\bibitem{DIBR_videoswebsite}
\BIBentryALTinterwordspacing
``{DIBR Videos Quality Database}.'' [Online]. Available:
  \url{http://ivc.univ-nantes.fr/en/databases/DIBR_Videos/} (Last visited Jan.
  2018).
\BIBentrySTDinterwordspacing

\bibitem{DIBR_imageswebsite}
\BIBentryALTinterwordspacing
``{DIBR Images Quality Database}.'' [Online]. Available:
  \url{http://ivc.univ-nantes.fr/en/databases/DIBR_Images/} (Last visited Jan.
  2018).
\BIBentrySTDinterwordspacing

\bibitem{FVSV_website}
\BIBentryALTinterwordspacing
``{Free-Viewpoint Synthesized Videos Quality Database }.'' [Online]. Available:
  \url{http://ivc.univ-nantes.fr/en/databases/Free-Viewpoint_synthesized_videos/}
  (Last visited Jan. 2018).
\BIBentrySTDinterwordspacing

\bibitem{SIAT_website}
\BIBentryALTinterwordspacing
``{SIAT Synthesized Video Quality Database}.'' [Online]. Available:
  \url{http://codec.siat.ac.cn/SIATDatabase/index.html} (Last visited Jan.
  2018).
\BIBentrySTDinterwordspacing

\bibitem{Song2015}
R.~Song, H.~Ko, and C.~C. Kuo, ``{MCL-3D: A database for stereoscopic image
  quality assessment using 2D-image-plus-depth source},'' \emph{Journal of
  Information Science and Engineering}, vol.~31, no.~5, pp. 1593--1611, Mar.
  2015.

\bibitem{MCL3D_website}
\BIBentryALTinterwordspacing
``{MCL 3D Database}.'' [Online]. Available:
  \url{http://mcl.usc.edu/mcl-3d-database/} (Last visited Jan. 2018).
\BIBentrySTDinterwordspacing

\bibitem{lafruit2015call}
G.~Lafruit, K.~Wegner, and M.~Tanimoto, ``Call for evidence on free-viewpoint
  television: Super-multiview and free navigation,'' \emph{MPEG N15348,
  Warsaw}, 2015.

\bibitem{lafruit2015draft}
------, ``Draft call for evidence on ftv,'' \emph{ISO/IEC JTC1/SC29/WG11
  MPEG2015 N}, vol. 15095, 2015.

\bibitem{NUS}
``Nagoya university sequences,''
  \url{http://www.fujii.nuee.nagoya-u.ac.jp/multiview-data/ }.

\bibitem{wegner34302ders}
K.~Wegner and O.~Stankiewicz, ``Ders software manual,'' \emph{ISO/IEC
  JTC1/SC29/WG11 M}, vol. 34302.

\bibitem{lafruit2015ftv}
G.~Lafruit, K.~Wegner, T.~Grajek, T.~Senoh, P.~Kov{\'a}cs, P.~Goorts,
  L.~Jorissen, B.~Ceulemans, P.~C. Lopez, S.~G. Lobo \emph{et~al.}, ``Ftv
  software framework,'' \emph{MPEG N15349, Warsaw}, 2015.

\bibitem{engelke2015perceived}
U.~Engelke and P.~Le~Callet, ``Perceived interest and overt visual attention in
  natural images,'' \emph{Signal Processing: Image Communication}, vol.~39, pp.
  386--404, 2015.

\bibitem{2engelke2015perceived}
------, ``Perceived interest and overt visual attention in natural images,''
  \emph{Signal Processing: Image Communication}, vol.~39, pp. 386--404, 2015.

\bibitem{itu2014subjective}
ITU, ``Methods for the subjective assessment of video quality, audio quality
  and audiovisual quality of internet video and distribution quality television
  in any environment,'' \emph{ITU-T Recommendation P.913}, 2014.

\bibitem{series2012methodology}
------, ``Methodology for the subjective assessment of the quality of
  television pictures,'' \emph{Recommendation ITU-R BT.500}, 2012.

\bibitem{VQEG2010b}
VQEG, ``{Report on the Validation of Video Quality Models for High Definition
  Video Content},'' Jun. 2010.

\bibitem{lim2013sketch}
J.~J. Lim, C.~L. Zitnick, and P.~Doll{\'a}r, ``Sketch tokens: A learned
  mid-level representation for contour and object detection,'' in
  \emph{Proceedings of the IEEE Conference on Computer Vision and Pattern
  Recognition}, 2013, pp. 3158--3165.

\bibitem{ling2017image}
S.~Ling and P.~Le~Callet, ``Image quality assessment for free viewpoint video
  based on mid-level contours feature,'' in \emph{IEEE International Conference
  on Multimedia and Expo}, Hong Kong, China, Jul. 2017, pp. 79--84.

\bibitem{ninassi2007does}
A.~Ninassi, O.~Le~Meur, P.~Le~Callet, and D.~Barba, ``Does where you gaze on an
  image affect your perception of quality? applying visual attention to image
  quality metric,'' in \emph{Image Processing, 2007. ICIP 2007. IEEE
  International Conference on}, vol.~2.\hskip 1em plus 0.5em minus 0.4em\relax
  IEEE, 2007, pp. II--169.

\bibitem{bay2008speeded}
H.~Bay, A.~Ess, T.~Tuytelaars, and L.~Van~Gool, ``Speeded-up robust features
  (surf),'' \emph{Computer vision and image understanding}, vol. 110, no.~3,
  pp. 346--359, 2008.

\bibitem{gu2017fast}
K.~Gu, L.~Li, H.~Lu, X.~Min, and W.~Lin, ``A fast reliable image quality
  predictor by fusing micro-and macro-structures,'' \emph{IEEE Transactions on
  Industrial Electronics}, vol.~64, no.~5, pp. 3903--3912, 2017.

\bibitem{liu2012image}
A.~Liu, W.~Lin, and M.~Narwaria, ``Image quality assessment based on gradient
  similarity,'' \emph{IEEE Transactions on Image Processing}, vol.~21, no.~4,
  pp. 1500--1512, 2012.

\bibitem{dollar2009integral}
P.~Doll{\'a}r, Z.~Tu, P.~Perona, and S.~Belongie, ``Integral channel
  features,'' 2009.

\bibitem{shechtman2007matching}
E.~Shechtman and M.~Irani, ``Matching local self-similarities across images and
  videos,'' in \emph{2007 IEEE Conference on Computer Vision and Pattern
  Recognition}.\hskip 1em plus 0.5em minus 0.4em\relax IEEE, 2007, pp. 1--8.

\bibitem{MSSIM}
Z.~Wang, E.~P. Simoncelli, and A.~C. Bovik, ``Multiscale structural similarity
  for image quality assessment,'' in \emph{Signals, Systems and Computers,
  2004. Conference Record of the Thirty-Seventh Asilomar Conference on},
  vol.~2.\hskip 1em plus 0.5em minus 0.4em\relax IEEE, 2003, pp. 1398--1402.

\bibitem{pinson2004new}
M.~H. Pinson and S.~Wolf, ``A new standardized method for objectively measuring
  video quality,'' \emph{IEEE Transactions on broadcasting}, vol.~50, no.~3,
  pp. 312--322, 2004.

\bibitem{VQM}
``Video quality metric (vqm) software,'' \url{http://www.
  its.bldrdoc.gov/resources/video-quality-research/software.aspx }.

\end{thebibliography}

\ifCLASSOPTIONcaptionsoff
  \newpage
\fi

%


%
%




\end{document}